\documentclass[conference]{IEEEtran}
\IEEEoverridecommandlockouts

\usepackage{times}
\usepackage[numbers]{natbib}
\usepackage{multicol}
\usepackage[bookmarks=true]{hyperref}
\usepackage{graphicx}
\graphicspath{{figures/}}
\usepackage{amsmath}
\usepackage{amssymb}
\usepackage{bm}
\usepackage{booktabs}
\usepackage{makecell}
\usepackage{multirow}
\usepackage{subfig}
\usepackage{xcolor}
\usepackage{pifont}
\usepackage[ruled,vlined]{algorithm2e}
\let\labelindent\relax
\usepackage[inline]{enumitem}
\usepackage{float}
\usepackage{stfloats}
\usepackage{tikz}

\newcommand{\cmark}{{\color{green!70!black}\ding{51}}}
\newcommand{\xmark}{{\color{red}\ding{55}}}

\begin{document}

\title{Efficient Feature-Free Initialization \\ for Monocular Visual-Inertial Systems Using a Feed-Forward 3D Model}

\author{
    Yuantai Zhang$^{1}$ \quad Jiaqi Yang$^{1}$ \quad Huajian Zeng$^{1}$ \quad Changhao Chen$^{2}$ \quad Haoang Li$^{2}$ \\
    Liang Li$^{3}$ \quad Dezhen Song$^{1}$ \quad Xingxing Zuo$^{1\dagger}$ \\[6pt]
    $^{1}$MBZUAI \quad $^{2}$HKUST (GZ) \quad $^{3}$Zhejiang University \\
    $^{\dagger}$Corresponding author
}

\maketitle

\begin{abstract}
Fast and reliable initialization is critical for monocular visual–inertial navigation systems (VINS), as it establishes the starting conditions for subsequent state estimation. Despite steady progress, most existing methods heavily rely on visual feature correspondences and require 3-4 seconds of sensory data for successful initialization, which limits their applicability and efficiency. With the advent of feed-forward 3D models that can directly predict point clouds from images, we revisit the visual–inertial initialization problem from a concise perspective. In this work, we propose a feature-free initialization framework that leverages up-to-scale point clouds predicted by a feed-forward 3D model, thereby obviating the need for visual feature tracking and estimation. This design substantially reduces system complexity and improves the reliability of initialization. 
Experiments on public datasets demonstrate that the proposed feature-free initialization method achieves the highest success rate, exceeding 90\%, and significantly reduces the data duration required for successful initialization, typically to under 1.2 s. We further validate our method on a self-collected dataset covering various indoor and outdoor scenarios, demonstrating robust performance, particularly in visually degraded environments where existing methods often fail. The code and dataset are available at \href{https://github.com/Yuantai-Z/FF-VIO-Init}{\textcolor{magenta}{\texttt{github.com/Yuantai-Z/FF-VIO-Init}}}.

%Real-world experiments on both publicly available datasets and our own collected dataset demonstrate that the proposed method can reliably initialize monocular VINS within 1.2 seconds, achieving fast and consistent performance across diverse scenarios. 

\end{abstract}

\section{Introduction}
Monocular visual-inertial navigation systems (VINS), which combine a single camera with an inertial measurement unit (IMU), enable robust and accurate pose estimation by leveraging the complementary characteristics of visual and inertial sensing. With continued advances in estimation theory, optimization techniques, and sensor fusion algorithms~\cite{Mourikis2007MultiStateConstraint, Leutenegger2014KeyframeBasedVisualInertial, Geneva2020OpenVINSResearch, Zuo2019LICFusionLiDARInertialCamera, Zhang2024moreprecise}, VINS has become a fundamental component in a wide range of applications, including augmented and virtual reality (AR/VR)~\cite{Yi2025EstimatingBody, Kong2025AriaGen}, robotics perception and navigation~\cite{Schmid2024Khronos, Kassab2024LanguageEXtendedIndoor, Zhang2024GNSSMultiSensor,Xu2025POGVINSTightly}.
A critical prerequisite for successful monocular VINS is reliable system initialization. Specifically, initialization aims to recover the initial linear velocity and gravity, as well as the metric scale that is inherently unobservable in visual-only pose estimation. Inaccurate initialization may introduce large estimation errors and may even lead to rapid divergence of the state estimator. At the same time, fast initialization is equally important, as it significantly expands the practical applicability of VINS in real-world scenarios such as on-device AR experiences, agile robotic deployment, and emergency navigation, where prolonged initialization periods are undesirable or infeasible.

Motivated by these requirements, numerous methods have been proposed to achieve fast and accurate initialization for monocular VINS~\cite{Dong-Si2012Estimatorinitialization, Zou2019StructVIOVisualInertial, Campos2021ORBSLAM3Accurate}. However, most existing approaches implicitly or explicitly rely on visual feature tracking across image views and accurate initialization of visual features (i.e., landmarks) from tracked 2D feature observations to establish geometric constraints between different camera frames. In practice, challenging conditions such as textureless environments, rapid camera motion, low-parallax trajectories, motion blur, poor illumination, and sensor noise frequently lead to poor feature tracking and unreliable landmark initialization. Under these conditions, existing monocular VINS initialization methods often become ill-conditioned, resulting in degraded accuracy, delayed convergence, or complete failure.

Recent advances in deep learning have catalyzed a paradigm shift in 3D geometric estimation. Feed-forward 3D models have demonstrated remarkable spatial understanding capabilities, enabling end-to-end prediction of scene-level 3D point clouds directly from input images~\cite{Wang2024DUSt3RGeometric, Wang2025VGGTVisuala, Wang2025Continuous3D}. However, such models typically suffer from inherent scale ambiguity. In this work, we explore the potential of leveraging feed-forward 3D models for fast and reliable monocular VINS initialization, while recovering the metric scale of the underlying 3D scene.

We propose a novel feature-free formulation for monocular VINS initialization that exploits geometric constraints from point clouds predicted by a feed-forward 3D model. 
Our approach bypasses nuisance visual feature extraction and initialization. The initialization pipeline consists of two stages: a closed-form linear system for an initial guess and a nonlinear visual-inertial bundle adjustment (VI-BA) for refinement. Our method not only simplifies the monocular VINS initialization problem into an extremely concise formulation, but also substantially improves its efficiency and robustness. The main contributions are:
\begin{itemize}
\item We propose an efficient monocular VINS initialization paradigm that leverages point cloud predictions from a feed-forward 3D model. 
To the best of our knowledge, this is the first work to integrate feed-forward 3D foundation models into monocular VINS initialization.
\item We derive a feature-free, closed-form linear system that naturally eliminates visual feature association while reliably estimating metric scale, initial velocity, and gravity. 
\item We formulate two variants of VI-BA leveraging constraints from the predicted point clouds: a feature-free formulation that achieves maximum efficiency and success rate, and a scale-constrained formulation that preserves feature correspondences for higher accuracy. 
\item We conduct extensive experiments on both public benchmarks and a self-collected dataset, achieving over 90\% success rate while reducing the required data duration across various challenging scenarios.   
\end{itemize}
\vspace{-2pt}
\begin{figure*}[t]
\centering
\includegraphics[width=1\textwidth]{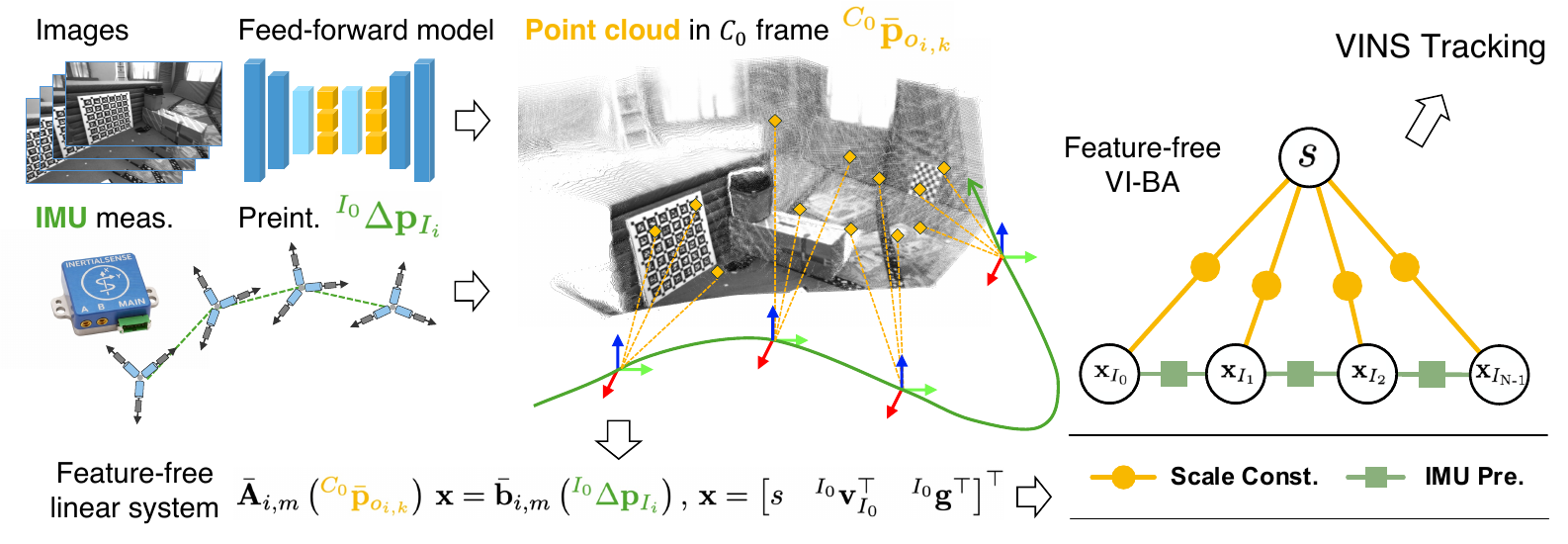}
\caption{System overview. A feed-forward 3D model predicts up-to-scale point clouds (yellow) from input images, while IMU measurements are preintegrated (green). These are combined into a feature-free closed-form linear system $\bar{\mathbf{A}}(\cdot)\mathbf{x}=\bar{\mathbf{b}}(\cdot)$, where $\bar{\mathbf{A}}(\cdot)$ incorporates up-to-scale point cloud geometry and $\bar{\mathbf{b}}(\cdot)$ encodes metric-scale IMU information. The closed-form solution is refined by a compact feature-free bundle adjustment involving only scale and IMU states, finally bootstrapping VINS tracking.}  
\label{fig:overview}
\vspace{-1em}
\end{figure*}

\section{Related Work}
\subsection{Monocular Visual-Inertial Initialization}
% Monocular visual-inertial initialization requires recovering several metric quantities, including gravity direction, initial velocity, and metric scale. Although these quantities can be obtained by assuming static motion, such assumptions are often violated in practice. Hence, most work has focused on dynamic initialization, which can be divided into tightly-coupled (closed-form) and loosely-coupled approaches.
Early initialization approaches often relied on static-motion assumptions, but such conditions are frequently violated in practice. Consequently, most existing work focuses on dynamic initialization, which can be broadly categorized into loosely coupled and tightly coupled (closed-form) approaches.
Loosely-coupled approaches rely on a successful visual odometry (VO) and align the up-to-scale camera trajectory with inertial constraints to recover metric scale. The initial states are recovered through different steps of linear least-squares~\cite{Qin2018VINSMonoRobust, Mur-Artal2017VisualInertialMonocular}. 
%Later work further improves this pipeline by accounting for the IMU probabilistic model and adding a quadratic constraint in nonlinear optimization form, respectively~\cite{Campos2020InertialOnlyOptimization, Zuniga-Noel2021AnalyticalSolution}. 
In contrast, closed-form solutions, pioneered by~\cite{Dong-Si2012Estimatorinitialization,Martinelli2014ClosedFormSolution}, tightly couple visual constraints and inertial measurements by jointly recovering feature positions together with the initial states. This approach is later extended by jointly estimating the gyroscope bias~\cite{Kaiser2017SimultaneousState}, along with additional observability tests to improve robustness~\cite{Campos2019FastRobust}. More recently, DRT~\cite{He2023RotationTranslationDecoupledSolution} and sqrtVINS~\cite{Peng2025sqrtmathbfVINS} exploited the coplanarity constraint of epipolar geometry to bypass the estimation for 3D features. However, all aforementioned methods fundamentally depend on reliable feature extraction and tracking, making them inherently sensitive to degeneracy and low-parallax scenarios.
\vspace{-2pt}
\subsection{Feed-Forward 3D Models}
Recent advances in 3D geometric perception have driven a paradigm shift from traditional iterative structure-from-motion (SfM) pipelines~\cite{Schonberger2016StructurefromMotionRevisited} to feed-forward models that directly infer scene geometry from images. 
DUSt3R~\cite{Wang2024DUSt3RGeometric} and MASt3R~\cite{Leroy2025GroundingImage} introduce end-to-end formulations that predict point clouds from image pairs, while VGGT~\cite{Wang2025VGGTVisuala} further generalizes this paradigm to multi-view inputs with strong cross-view consistency. 
Follow-up efforts extend feed-forward 3D models toward longer sequence settings using recurrent network architecture \cite{Wang2025Continuous3D,Chen2025TTT3R3D}, spatial memory~\cite{wang20243d}, and SLAM pipeline~\cite{Murai2025MASt3RSLAMRealTime,Maggio2025VGGTSLAMDense}. Despite these advances, feed-forward 3D models typically recover geometry only up to an unknown similarity transformation~\cite{Wang2024DUSt3RGeometric,Wang2025VGGTVisuala}.
Several efforts address this limitation by predicting metric depth~\cite{hu2024metric3d} or incorporating explicit scale regression with optional geometric priors~\cite{Keetha2025MapAnythingUniversal, wang2025amb3r}. 
However, in large-scale or long-horizon scenarios, visual metric-scale estimation remains prone to scale drift and instability~\cite{Murai2025MASt3RSLAMRealTime}. 
% As a result, metric scale, gravity direction, and inertial states remain insufficiently constrained, limiting the applicability of these methods to visual-inertial navigation.
\vspace{-3pt}
\subsection{Learning-aided Visual-Inertial Initialization}

Deep learning has been widely integrated in visual SLAM and VINS to enhance robustness and accuracy. These approaches either replace core SLAM components with learning-based counterparts~\cite{Teed2021DROIDSLAMDeep, Wang2025Imperativelearning,Yuan2025SLAMFormerPutting}, or augment traditional pipelines with learned priors~\cite{Zuo2021CodeVIOVisualInertialOdometry,Merrill2024VisualInertialSLAM}. Our work falls into the latter category.
In particular, the emergence of 3D foundation models has enabled SLAM systems to leverage their strong spatial prediction. MASt3R-SLAM~\cite{Murai2025MASt3RSLAMRealTime} and VGGT-SLAM~\cite{Maggio2025VGGTSLAMDense} leverage dense predictions for tracking and mapping, later extending to multi-sensor settings with IMU integration~\cite{Wang2025LiDARVGGTCrossModal}. However, these systems assume a valid initial state is already available. To the best of our knowledge, exploiting feed-forward 3D geometry for visual–inertial initialization has not yet been explored.
The works most closely related to ours are~\cite{Zhou2022LearnedMonocular,Merrill2023FastMonocular}, which leverage learned single-view depth for monocular VINS initialization.
Zhou et al.~\cite{Zhou2022LearnedMonocular} incorporate depth predictions as additional measurements in visual-inertial bundle adjustment, while Merrill et al.~\cite{Merrill2023FastMonocular} derive a closed-form solution that eliminates explicit 3D feature states.
These works demonstrate that learned geometric priors can improve initialization robustness, yet they still rely on feature correspondences.

\section{Visual-Inertial Initialization}
This section presents the measurement models for IMU and camera, and revisits the standard monocular visual-inertial initialization pipeline consisting of a closed-form linear system and nonlinear refinement.
\subsection{IMU and Camera Models}
An IMU measures angular velocity $\bm{\omega}$ and acceleration $\bm{a}$, modeled as:
\begin{align}
  \tilde{\bm{\omega}} &= \bm{\omega} +\mathbf{b}_\omega+ \mathbf{n}_\omega \\
  \tilde{\bm{a}} &= \bm{a} + {}^{I}_{G}\mathbf{R} {}^{G}\mathbf{g} + \mathbf{b}_a + \mathbf{n}_a
\end{align}
where $\tilde{\bm{\omega}}$ and $\tilde{\bm{a}}$ are the measured angular velocity and acceleration, respectively. ${}^{I}_{G}\mathbf{R}$ is the rotation to transform a point in gravity-aligned global frame $\{\bm{G}\}$ to IMU frame $\{\bm{I}\}$, $ {}^{G}\mathbf{g} = [0,0,g]^\top $ is the gravity vector in global frame with $g = 9.81\,\mathrm{m/s^2}$.  Based on classical inertial navigation system (INS) kinematic model~\cite{Savage1998StrapdownInertiala}, the discrete-time propagation from frame $i$ to $i+1$ is:
\begin{align}
{}^{G}_{I_{i+1}}\mathbf{R} &= {}^{G}_{I_i}\mathbf{R}\, {}^{I_i}_{I_{i+1}}\Delta \mathbf{R}\\
{}^{G}\mathbf{v}_{I_{i+1}} &= {}^{G}\mathbf{v}_{I_i} - {}^{G}\mathbf{g}\Delta t + {}^{I_i}_{G}\mathbf{R}^\top {}^{I_i}\Delta \mathbf{v}_{I_{i+1}} \\
{}^{G}\mathbf{p}_{I_{i+1}} &= {}^{G}\mathbf{p}_{I_i} + {}^{G}\mathbf{v}_{I_i}\Delta t - \frac{1}{2}{}^{G}\mathbf{g}\Delta t^2 + {}^{I_i}_{G}\mathbf{R}^\top {}^{I_i}\Delta \mathbf{p}_{I_{i+1}} \label{eq:imu-pro-pos}
\end{align}
where $\{{}^{G}_{I_i}\mathbf{R},\, {}^{G}\mathbf{p}_{I_i}\}$ denotes the IMU pose in $\{\bm{G}\}$ at timestamp $i$, together forming the rigid-body transformation ${}^{G}_{I_i}\mathbf{T} \in SE(3)$. $\mathbf{z}_{i,i+1}^{\mathcal{I}} = \{\Delta \mathbf{R}(\tilde{\bm{\omega}}), \Delta \mathbf{v}(\tilde{\bm{\omega}},\tilde{\bm{a}}), \Delta \mathbf{p}(\tilde{\bm{\omega}},\tilde{\bm{a}})\}$ represents the IMU preintegration measurements between consecutive frames~\cite{Lupton2012VisualInertialAidedNavigation, Forster2015IMUPreintegration}, where $\Delta \mathbf{R}$, $\Delta \mathbf{v}$, and $\Delta \mathbf{p}$ denote the preintegrated relative rotation, velocity, and position, respectively. 
% , where:
% \begin{align}
% {}^{I_{i}}_{I_{i+1}}\Delta \mathbf{R} &= \int_{t_i}^{t_{i+1}} \tilde{\bm{\omega}}(t_\tau)\, d\tau  \notag \\
% {}^{I_i}\Delta \mathbf{v}_{I_{i+1}} &= \int_{t_i}^{t_{i+1}} {}^{I_{i}}_{I_\tau}\Delta \mathbf{R}\,\big(\tilde{\bm{a}}(\tau )-\mathbf{b}_a(\tau )-\mathbf{n}_a(\tau)\big)\, d\tau \notag \\
% {}^{I_i}\Delta \mathbf{p}_{I_{i+1}} &= \iint_{t_i}^{t_{i+1}}  {}^{I_{i}}_{I_\tau}\Delta \mathbf{R}\,\big(\tilde{\bm{a}}(\tau )-\mathbf{b}_a(\tau )-\mathbf{n}_a(\tau)\big)\, d\tau^2  \notag
% \end{align}
In the closed-form system, we use the first IMU frame $\{\bm{I_0}\}$ as the global frame.
% , where the IMU kinematic model can be further simplified as following:
% \begin{align}
% {}^{I_0}_{I_i}\mathbf{R} & = {}^{I_0}_{I_i}\Delta \mathbf{R} \\
% {}^{I_0}\mathbf{v}_{I_i} & = {}^{I_0}\mathbf{v}_{I_0} - {}^{I_0}\mathbf{g}\,\Delta t + {}^{I_0}\Delta \mathbf{v}_{I_i} \\
% {}^{I_0}\mathbf{p}_{I_i} & = {}^{I_0}\mathbf{v}_{I_0}\Delta t - \frac{1}{2}\,{}^{I_0}\mathbf{g}\,\Delta t^{2} + {}^{I_0}\Delta \mathbf{p}_{I_i}
% \label{eq:imu-pro-pos-I0}
% \end{align}

% Begin introducing camera model, this should include both
For the camera model, we consider a monocular pinhole camera rigidly attached and temporally synchronized to the IMU, with known camera intrinsics and extrinsics $\{{}^{C}_{I}\mathbf{R},{}^{C}\mathbf{p}_I\}$.  The observation of the $m$-th visual feature is denoted by $\mathbf{z}_{i,m}^{\mathcal{C}} \in \mathbb{R}^2$, and the measurement model can be written as:
\begin{align}
\mathbf{z}_{i,m}^{\mathcal{C}}
=&
\begin{bmatrix}
u_{i,m} \\
v_{i,m}
\end{bmatrix}
=
\pi\left({}^{C_i}\mathbf{p}_{f_m}\right)
+
\mathbf{n}
\label{eq:visual_measurement_model} \\
{}^{C_i}\mathbf{p}_{f_m}
=&
{}^{C}_{I}\mathbf{R}\;
{}^{I_i}_{I_0}\mathbf{R}\,
\big(
{}^{I_0}\mathbf{p}_{f_m} - {}^{I_0}\mathbf{p}_{I_i}
\big)
+ {}^{C}\mathbf{p}_I
\label{eq:basic_reprojection_model}
\end{align}
where $(u_{i,m}, v_{i,m})$ denotes the normalized coordinates of the $m$-th feature at time $t_i$, $\pi(\cdot)$ is the normalized projection function, $\{\bm{C}_i\}$ is the camera frame at time $t_i$, and ${}^{I_0}\mathbf{p}_{f_m}$ is the 3D position of the $m$-th feature expressed in frame $\{\bm{I}_0\}$.
%, and $\{{}^{I_0}_{I_i}\mathbf{R}, {}^{I_0}\mathbf{p}_{I_i}\}$ represents the IMU pose.
Eq.~\eqref{eq:basic_reprojection_model} can be rewritten as the following linear constraint~\cite{Dong-Si2012Estimatorinitialization}:
\begin{align}
\begin{bmatrix}
1 & 0 & -u_{i,m} \\
0 & 1 & -v_{i,m}
\end{bmatrix}
\left(
{}^{C}_{I}\mathbf{R}\;
{}^{I_i}_{I_0}\mathbf{R}\,
\big(
{}^{I_0}\mathbf{p}_{f_m} - {}^{I_0}\mathbf{p}_{I_i}
\big)
+ {}^{C}\mathbf{p}_I
\right) = 0
\label{eq:basic_reprojection_model_linear}
\end{align}
\subsection{VI Initialization}
The minimal goal of VINS initialization is to recover the initial velocity ${}^{I_0}\mathbf{v}_{I_0}$ and gravity vector ${}^{I_0}\mathbf{g}$. Typically, two phases are involved: (i) a linear system that provides a coarse initial guess of the state; (ii) a nonlinear optimization that refines the estimates and outputs all necessary states to bootstrap the pose tracking of VINS systems.
\subsubsection{Constrained Linear Least-Squares Formulation}
By substituting Eq.~\eqref{eq:imu-pro-pos} into Eq.~\eqref{eq:basic_reprojection_model_linear} and stacking all visual measurements, we obtain a linear system $\mathbf{A}\mathbf{x} = \mathbf{b}$. The state vector $\mathbf{x}$ contains all $M$ feature positions, initial velocity, and gravity:
% \begin{align}
%     &\begin{bmatrix}
%     1 & 0 & -u_{i,m} \\
%     0 & 1 & -v_{i,m}
%     \end{bmatrix}
%     \Bigg[ {}^{C}_{I}\mathbf{R}\, {}^{I_i}_{I_0}\mathbf{R} \notag \\
%     &\quad \left( {}^{I_0}\mathbf{p}_{f_m}
%     - {}^{I_0}\mathbf{v}_{I_0}\Delta t
%     + \frac{1}{2}{}^{I_0}\mathbf{g}\,\Delta t^2
%     - \Delta{}^{I_0}\mathbf{p}_{I_i} \right)
%     + {}^{C}\mathbf{p}_{I} \Bigg] = \bm{0}
% \end{align}
\begin{equation}
\mathbf{x} =
\begin{bmatrix}
{}^{I_0}\mathbf{p}_{f_1}^\top & \cdots & {}^{I_0}\mathbf{p}_{f_M}^\top &
{}^{I_0}\mathbf{v}_{I_0}^\top &
{}^{I_0}\mathbf{g}^\top
\end{bmatrix}^\top
\label{eq:linear_state_vector_feature}
\end{equation}
Each measurement $\mathbf{z}_{i,m}^{\mathcal{C}}$ contributes two rows to $\mathbf{A}$ and $\mathbf{b}$:
\begin{align}
\mathbf{A}_{i,m} &= \mathbf{H}_{i,m}
\begin{bmatrix}
\mathbf{E}_m & -\Delta t\,\mathbf{I}_3 & \frac{1}{2}\Delta t^2\,\mathbf{I}_3
\end{bmatrix} \notag \\
\mathbf{b}_{i,m} &= \mathbf{H}_{i,m}\left({}^{I_0}\Delta \mathbf{p}_{I_i}
- {}^{I_0}_{I_i}\mathbf{R}\,{}^{I}_{C}\mathbf{R}\,{}^{C}\mathbf{p}_{I} \right) 
\label{eq:linear_feature}
\end{align}
where $\mathbf{E}_m=[\mathbf{0}_{3\times 3(m-1)}\ \mathbf{I}_3\ \mathbf{0}_{3\times 3(M-m)}]$ selects the $m$-th feature position from $\mathbf{x}$, and $\mathbf{H}_{i,m}$ is defined as:
\begin{align}
\mathbf{H}_{i,m} =
\begin{bmatrix}
1 & 0 & -u_{i,m} \\
0 & 1 & -v_{i,m}
\end{bmatrix}
{}^{C}_{I}\mathbf{R}\,{}^{I_i}_{I_0}\mathbf{R}
\label{eq:H_matrix}
\end{align}
The linear system can be solved as a constrained least-squares problem subject to $\lVert {}^{I_0}\mathbf{g} \rVert^2_2 = g^2$~\cite{Dong-Si2012Estimatorinitialization}. Once the gravity vector ${}^{I_0}\mathbf{g}$ is estimated, the gravity-aligned rotation ${}^{G}_{I_0}\mathbf{R}$ can be recovered via Gram-Schmidt orthogonalization. 
% The translation ${}^{G}\mathbf{p}_{I_0} = \mathbf{0}$ since the origins of $\{\bm{G}\}$ and $\{\bm{I_0}\}$ coincide.

\subsubsection{Nonlinear Refinement}
The linear solution provides a coarse estimate that may be affected by noise. To properly account for measurement uncertainties, it is usually refined through nonlinear optimization. Given $N$ keyframes in the initialization window, the full state vector is defined as:
\begin{align}
\mathcal{X} &=
\big[
\mathbf{x}_{I_0}^\top \ \cdots \ \mathbf{x}_{I_{N-1}}^\top \
{}^{G}\mathbf{p}_{f_1}^\top \ \cdots \ {}^{G}\mathbf{p}_{f_M}^\top
\big]^\top \label{eq:nonlinear_system_feature_statevector}\\
\mathbf{x}_{I_i} &=
\big[
{}^{I_i}_{G}\mathbf{q}^\top \
{}^{G}\mathbf{p}_{I_i}^\top \
{}^{G}\mathbf{v}_{I_i}^\top \
\mathbf{b}_{\omega}^\top \
\mathbf{b}_{a}^\top
\big]^\top
\end{align}
where $\mathbf{x}_{I_i}$ denotes the 15-dimensional IMU state that contains orientation, velocity, position, and biases of the $i$-th frame.

The optimization minimizes a cost function comprising three terms, as illustrated by the factor graph in Fig.~\ref{fig:nonlinear_optimization}(a):
\begin{align}
\mathcal{X}^* &= \arg\min_{\mathcal{X}} \Big\{
\sum_{i \in \mathcal{I}} \left\lVert \mathbf{r}_{\mathcal{I}}(\mathbf{z}_{i,i+1}^{\mathcal{I}}, \mathbf{x}_{I_i}, \mathbf{x}_{I_{i+1}}) \right\rVert^{2}_{\boldsymbol{\Sigma}_{\mathcal{I}}} \notag \\
&+ \sum_{(i,m) \in \mathcal{C}} \left\lVert \mathbf{r}_{\mathcal{C}}(\mathbf{z}_{i,m}^{\mathcal{C}}, \mathbf{x}_{I_i}, \mathbf{p}_{f_m}) \right\rVert^{2}_{\boldsymbol{\Sigma}_{\mathcal{C}}} + \left\lVert \mathbf{r}_{\mathrm{pr}}(\mathbf{z}^{\mathrm{p}}, \mathbf{x}_{I_0}) \right\rVert^{2}_{\boldsymbol{\Sigma}_{\mathrm{pr}}}
\Big\}
\label{eq:nonlinear_system_feature}
\end{align}
where $\mathbf{r}_{\mathcal{C}}$ is the reprojection residual as in Eq.~\eqref{eq:visual_measurement_model}, $\mathbf{r}_{\mathcal{I}}$ is the IMU preintegration residual~\cite{Lupton2012VisualInertialAidedNavigation, Forster2015IMUPreintegration}, and $\mathbf{r}_{\mathrm{pr}}$ is the prior residual that constrains the state to the linearization point. 
$\boldsymbol{\Sigma}_{\mathcal{I}}$ is propagated from IMU noise during preintegration, $\boldsymbol{\Sigma}_{\mathcal{C}}$ is set according to pixel measurement noise, and $\boldsymbol{\Sigma}_{\mathrm{pr}}$ is manually configured with small variance on unobservable states to stabilize optimization~\cite{Hesch2014ConsistencyAnalysis}.

\section{Feature-Free Initialization}
The classical pipeline described above suffers from two fundamental limitations: (i) feature extraction and tracking are fragile under degraded scenes; (ii) jointly estimating a large number of feature positions inflates the state dimension and degrades numerical conditioning. We address both issues by replacing feature correspondences with up-to-scale point clouds from a feed-forward 3D model as illustrated in Fig.~\ref{fig:overview}, yielding a feature-free initialization framework with a compact state space and improved efficiency.
\subsection{Feed-Forward 3D Models}
We incorporate feed-forward 3D foundation models within the classical VINS initialization framework for faster and more robust state recovery. Previous works leverage monocular depth prediction networks~\cite{Zhou2022LearnedMonocular,Merrill2023FastMonocular}, which provide 3D geometry expressed in the local camera frame of each image. These methods implicitly rely on feature correspondences to establish geometric constraints across multiple frames, limiting their applicability. Crucially, 3D foundation models produce point clouds expressed in a consistent reference camera frame $\{\bm{C}_0\}$. This property allows us to bypass the need for the whole feature extraction and tracking process, and to directly use geometry from the point cloud for state estimation.

We employ the pretrained VGGT~\cite{Wang2025VGGTVisuala} as our feed-forward 3D model $\mathcal{F}$ to infer point clouds.
% As shown in Fig.~\ref{fig:vggtinfer},
Given $N$ input images $\{\mathcal{I}_i\}_{i=0}^{N-1}$ of size $H \times W$, the model outputs:
\begin{equation}
\mathcal{F}\left(\{\mathcal{I}_i\}_{i=0}^{N-1}\right) = \left\{ {}^{C_0}\bar{\mathbf{p}}_{o_{i,k}},\, c_{i,k} \right\}^{H \times W \times N}_{i \in N,\, k \in H \times W}
\label{eq:vggt_output}
\end{equation}
where ${}^{C_0}\bar{\mathbf{p}}_{o_{i,k}} \in \mathbb{R}^3$ is the up-to-scale 3D position of the $k$-th point from image $\mathcal{I}_i$, with subscript $o$ denoting per-pixel point cloud rather than tracked features. $c_{i,k} \in [1,\infty]$ is the confidence score indicating the prediction uncertainty. To keep the problem tractable, we do not use dense prediction. Instead, we first filter out low-confidence predictions according to $c_{i,k}$. We then partition each image into uniform patches and sample a fixed number $K$ of high-confidence points per frame.
\subsection{Robust Feature-Free Closed-Form Solution}
In SLAM systems, 3D points are typically expressed by 3D coordinates, depth, or inverse depth~\cite{Civera2008InverseDepth}. In this work, given the predicted up-to-scale point cloud $\left\{{}^{C_0}\bar{\mathbf{p}}_{o_{i,k}} \right\}$, we represent 3D points by a scale parameter $s$ as:
%$\left\{ {}^{C_0}\bar{\mathbf{p}}_{o_{i,k}} \right\}^{KN}_{i \in N,\, k \in K}$
\begin{align}
    {}^{I_0}\mathbf{p}_{o_{i,k}} = s \, {}^{I}_{C}\mathbf{R} \, {}^{C_0}\bar{\mathbf{p}}_{o_{i,k}} + {}^{I}\mathbf{p}_{C} \label{eq:scale-the-point}
\end{align}

Any pixel of $\{\mathcal{I}_i\}_{i=0}^{N-1}$ can be mapped to a 3D point ${}^{I_0}\mathbf{p}_{o_{i,k}}$, and all points share a single unknown scale factor $s$. Unlike traditional methods that estimate $M$ features and require a $3M$-dimensional feature state (see Eq.~\eqref{eq:linear_state_vector_feature}), we reduce the feature-related state to a single scalar, significantly lowering the problem dimensionality. Substituting Eq.~\eqref{eq:scale-the-point} into Eq.~\eqref{eq:basic_reprojection_model} and stacking all measurements yields a linear system $\bar{\mathbf{A}}\mathbf{x} = \bar{\mathbf{b}}$, where the state vector contains only scale, initial velocity, and gravity:
% \begin{align}
%     {}^{C_i}\mathbf{p}_{f_m}
%     &= {}^{C}_{I}\mathbf{R} \, {}^{I_i}_{I_0}\mathbf{R}
%        \left( s \, {}^{I}_{C}\mathbf{R} \, {}^{C_0}\bar{\mathbf{p}}_{f_m} + {}^{I}\mathbf{p}_{C} - {}^{I_0}\mathbf{p}_{I_i} \right)
%        + {}^{C}\mathbf{p}_{I} \label{eq:pointcloud-lls}
% \end{align}
\begin{equation}
\mathbf{x} =
\begin{bmatrix}
s & {}^{I_0}\mathbf{v}_{I_0}^\top & {}^{I_0}\mathbf{g}^\top
\end{bmatrix}^\top
\label{eq:linear_state_vector_scale}
\end{equation}
Each measurement $\mathbf{z}_{i,k}^{\mathcal{C}}$ contributes two rows to $\bar{\mathbf{A}}$ and $\bar{\mathbf{b}}$:
\begin{align}
\bar{\mathbf{A}}_{i,k} &= \mathbf{H}_{i,k}
\begin{bmatrix}
{}^{I}_{C}\mathbf{R}\, {}^{C_0}\bar{\mathbf{p}}_{o_{i,k}} & -\Delta t\,\mathbf{I}_3 & \frac{1}{2}\Delta t^2\,\mathbf{I}_3
\end{bmatrix} \notag \\
\bar{\mathbf{b}}_{i,k} &= \mathbf{H}_{i,k}\left({}^{I_0}\Delta \mathbf{p}_{I_i} - {}^{I_0}_{I_i}\mathbf{R}\,{}^{I}_{C}\mathbf{R}\,{}^{C}\mathbf{p}_{I} + {}^{I}_{C}\mathbf{R}\,{}^{C}\mathbf{p}_{I}\right) \label{eq:linear_system_scale}
\end{align}

From the $K$ sampled points from each of the $N$ frames, we form a linear system with $\bar{\mathbf{A}} \in \mathbb{R}^{2KN \times 7}$ and $\bar{\mathbf{b}} \in \mathbb{R}^{2KN}$, and solve $\mathbf{x} \in \mathbb{R}^{7}$ via weighted constrained least-squares, where the predicted confidence $c_{i,k}$ is used to suppress unreliable predictions.
Due to the fixed state dimension, RANSAC can be directly integrated into the solving process to reject outliers in the predicted point cloud and improve robustness~\cite{Merrill2023FastMonocular}. The RANSAC procedure is provided in supplementary material.

\subsection{Nonlinear Refinement}
\begin{figure*}[!htbp]
\centering
\includegraphics[width=0.95\textwidth]{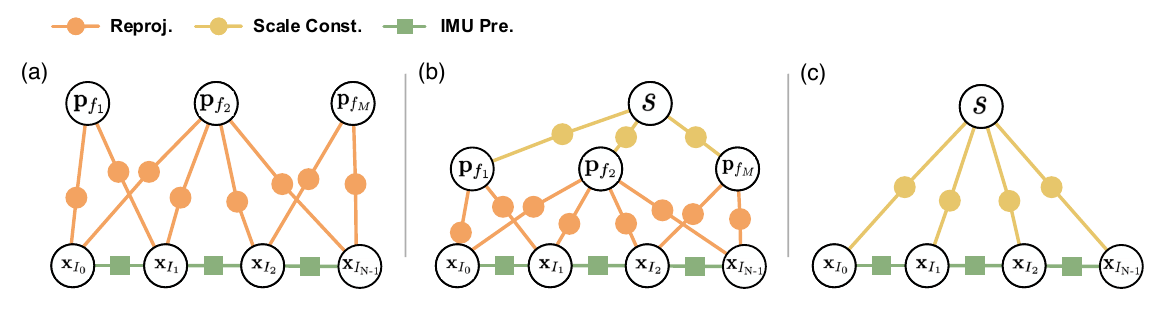}
\caption{Comparison of factor graphs for nonlinear refinement. (a) Traditional feature-based optimization jointly optimizes IMU states and 3D feature positions (see Eq.~\ref{eq:nonlinear_system_feature}). (b) Our scale-constrained formulation augments feature-based optimization with shared scale factors (see Eq.~\eqref{eq:nonlinear_system_feature_scale}). (c) Our feature-free formulation eliminates features, only containing IMU states and scales.}
\label{fig:nonlinear_optimization}
\vspace{-1em}
\end{figure*}
We propose two VI-BA formulations that leverage the predicted point cloud: (1) A Scale-Constrained (SC) formulation that augments traditional feature-based optimization with scale factors, preserving feature correspondences for higher accuracy; and (2) A Feature-Free (FF) formulation that eliminates feature states entirely, achieving maximum efficiency and robustness.
Both formulations directly reuse the point cloud from the linear system, and thus directly benefit from the RANSAC inlier set.
\subsubsection{Scale-Constrained Formulation} As shown in Fig.~\ref{fig:nonlinear_optimization}(b), additional scale parameters $\mathbf{s}$ are introduced into the feature-based BA states (see Eq.~\eqref{eq:nonlinear_system_feature_statevector}). The VI-BA problem is formulated as:
% The state vector is defined as:
% \begin{align}
% \mathcal{X} &=
% \big[
% \mathbf{x}_{I_0}^\top \ \cdots \ \mathbf{x}_{I_N}^\top \
% {}^{G}\mathbf{p}_{f_1}^\top \ \cdots \ {}^{G}\mathbf{p}_{f_M}^\top \
% \mathbf{s}^\top
% \big]^\top
% \end{align}
% where $\mathbf{s}$ contains multiple scale parameters.
\begin{align}
\mathcal{X}^* = \arg\min_{\mathcal{X}} \Big\{
&{\textstyle\sum} \| \mathbf{r}_{\mathcal{I}} \|^{2}_{\boldsymbol{\Sigma}_{\mathcal{I}}}
+ {\textstyle\sum} \| \mathbf{r}_{\mathcal{C}} \|^{2}_{\boldsymbol{\Sigma}_{\mathcal{C}}} \notag \\
&+ {\textstyle\sum} \| \mathbf{r}_{\bar{\mathbf{p}}} \|^{2}_{\boldsymbol{\Sigma}_{\bar{\mathbf{p}}}}
+ \| \mathbf{r}_{\mathrm{pr}} \|^{2}_{\boldsymbol{\Sigma}_{\mathrm{pr}}}
\Big\}
\label{eq:nonlinear_system_feature_scale}
\end{align}
where $\mathbf{r}_{\bar{\mathbf{p}}}$ is the scale residual that connects multiple features to the shared scale parameter $s$, providing additional geometric constraints from the predicted point cloud:
\begin{align}
\mathbf{r}_{\bar{\mathbf{p}}}({}^{C_0}\bar{\mathbf{p}}_{f_m},{}^{G}\mathbf{p}_{f_m}, s) &= {}^{G}\mathbf{p}_{f_m} - (s \, {}^{G}_{C_0}\mathbf{R} \, {}^{C_0}\bar{\mathbf{p}}_{f_m} + {}^{G}\mathbf{p}_{C_0})
\end{align}
Here ${}^{C_0}\bar{\mathbf{p}}_{f_m}$ is obtained by bilinearly interpolating $\{{}^{C_0}\bar{\mathbf{p}}_{o_{i,k}}\}$ in each frame at the tracked feature location. The measurement covariance ${\boldsymbol{\Sigma}_{\bar{\mathbf{p}}}}$ is determined by the predicted confidence score.
\subsubsection{Feature-Free Formulation}
In the second approach, we take a step further to completely eliminate features by estimating 3D points solely through the scale factor $s$, as shown in Fig.~\ref{fig:nonlinear_optimization}(c). The state vector of feature-free nonlinear optimization is defined as:
\begin{align}
\mathcal{X} &= \big[ \mathbf{x}_{I_0}^\top \ \cdots \ \mathbf{x}_{I_{N-1}}^\top \ \mathbf{s}^\top \big]^\top
\end{align}

The residuals contained in our feature-free VI-BA are the same as Eq.~\eqref{eq:nonlinear_system_feature}, however, the reprojection residual $\mathbf{r}_{\mathcal{C}}$ is parameterized by the scale factor $s$ instead of the 3D point. For clarity, we provide the explicit form:
\begin{align}
\mathbf{r}_{\mathcal{C}}({}^{C_0}\bar{\mathbf{p}}_{o_{i,k}}, \mathbf{x}_{I_i}, s) = \mathbf{z}_{i,k}^{\mathcal{C}} - \pi\big({}^{C}_{I}\mathbf{T} \, {}^{I_i}_{G}\mathbf{T} \, {}^{G}_{I_0}\mathbf{T} \, {}^{I}_{C}\mathbf{T} \, (s \, {}^{C_0}\bar{\mathbf{p}}_{o_{i,k}})\big)
\label{eq:reproj_feature_free}
\end{align}
The detailed expansion of $\mathbf{r}_{\mathcal{C}}$ together with its Jacobians w.r.t.\ the scale factor and IMU states is provided in Sec.~VII-C of the supplementary material.
We emphasize that $\mathbf{z}_{i,k}^{\mathcal{C}}$ can be any pixel coordinates in image $\mathcal{I}_i$, not restricted to feature correspondences. The shared index $k$ is used solely for bookkeeping of sampled points within each frame and does not imply data association across views. 
Since all points share the same scale factor $s$, this formulation significantly reduces the state dimension from $\mathcal{O}(3M)$ of Eq.~\eqref{eq:nonlinear_system_feature} to $\mathcal{O}(1)$ for feature variables, allowing us to only estimate scales and IMU states.

By coupling this feature-free VI-BA (Eq.~\eqref{eq:reproj_feature_free}) with the feature-free closed-form linear system (Eq.~\eqref{eq:linear_system_scale}), we obtain a fully feature-free VI initialization pipeline, denoted \textbf{Ours~(FF)}, in contrast to the scale-constrained counterpart \textbf{Ours~(SC)} that retains feature-based BA\@. This unified FF pipeline offers several advantages:
\begin{itemize}
  \item \textbf{Efficiency.}
  The compact formulation leads to lower computational complexity and fast convergence.
  \item \textbf{Robustness.}
  Being feature-free, the pipeline is inherently immune to common visual issues such as feature noise, mismatches, and insufficient parallax.
  \item \textbf{Numerical stability.}
  The largely reduced state dimension leads to better-conditioned BA.
\end{itemize}

\textbf{Remark:}
Directly optimizing scale can cause degeneracy under measurement noise, where the scale tends to collapse toward zero. We address this by reparameterizing the scale as~\cite{Garg2019LearningSingle}:
\begin{align}
s = \varepsilon + \log(e^{\tilde{s}} + 1)
\end{align}
where $\varepsilon = 10^{-5}$. This softplus-based formulation ensures $s > 0$ while enabling unconstrained optimization of $\tilde{s}$.

Moreover, although theoretically a single scale parameter suffices for the entire scene, in practice, prediction errors may vary across different regions. Thus, in nonlinear refinement, we partition the point cloud into a 2D grid along the principal plane, and assign an independent scale parameter to each region to absorb local errors. To prevent discontinuities, neighboring regions are connected by a smoothness constraint:
\begin{align}
\mathbf{r}_{\Delta \mathrm{s}}(s_j, s_l) &= s_j - s_l
\end{align}
where $s_j$ and $s_l$ are scale parameters of adjacent regions. In our implementation, we use a $3 \times 3$ grid, resulting in only 9 scale parameters. The effect of regional scales is evaluated in the ablation study.

\section{Experiments}
\begin{table*}[!htbp]
\centering
\caption{Evaluation of linear systems on EuRoC dataset: gravity error ($^\circ$) $\downarrow$ / velocity error (m/s) $\downarrow$.}
% \small
\vspace{-0.5em}
\label{tab:euroc_linear}
\begin{tabular*}{\textwidth}{@{\extracolsep{\fill}}lccccccc}
\toprule
Algorithm & V1\_01 & V1\_02 & V1\_03 & V2\_01 & V2\_02 & V2\_03 & Average \\
\midrule
Dong-Si & 2.42 / 0.43 & 10.29 / 0.75 & 7.54 / 0.79 & 2.53 / 0.33 & 7.41 / 0.58 & 7.11 / 0.63 & 6.22 / 0.59 \\
Dep.(AB) & 2.37 / 0.40 & 4.60 / 0.85 & 5.33 / 0.85 & 1.85 / 0.48 & 4.15 / 0.60 & 4.73 / 0.75 & 3.84 / 0.66 \\
sqrtVINS & 2.59 / 0.24 & 3.88 / 0.63 & 5.42 / 0.50 & 4.39 / 0.29 & 6.60 / 0.51 & 1.56 / 0.21 & 4.07 / \textbf{0.40} \\
DRT & 2.28 / 0.29 & 4.10 / 0.60 & 6.68 / 0.52 & 1.91 / 0.21 & 3.83 / 0.31 & 3.09 / 0.51 & 3.65 / 0.41 \\
Ours w/o RANSAC & 2.05 / 0.37 & 3.12 / 0.82 & 6.37 / 0.74 & 2.02 / 0.35 & 4.42 / 0.63 & 3.47 / 0.54 & 3.58 / 0.58 \\
Ours & 2.27 / 0.40 & 3.17 / 0.81 & 6.11 / 0.70 & 2.01 / 0.35 & 3.99 / 0.47 & 3.71 / 0.65 & \textbf{3.54} / 0.56 \\
\bottomrule
\end{tabular*}
\vspace{-0.5em}
\end{table*}

\begin{table*}[!htbp]
\centering
\caption{Overall performance on EuRoC dataset (Setting: 5 keyframes, 0.5\,s window).}
\vspace{-0.5em}
\label{tab:euroc_summary}
\setlength{\tabcolsep}{10pt}
\begin{tabular}{lccccccc}
\toprule
 & Dong-Si & Dep.(AB) & Dep.(Const.) & sqrtVINS & DRT & Ours~(SC) & Ours~(FF) \\
\midrule
Win. ATE $\downarrow$ & 1.47/0.024 & 1.25/\textbf{0.020} & \textbf{1.20}/0.026 & \textbf{1.20}/0.022 & 1.52/0.031 & \textbf{1.20/0.020} & 3.53/0.034 \\
Traj. ATE $\downarrow$ & 4.30/0.078 & 4.07/0.087 & 4.01/0.079 & 3.48/0.082 & 4.44/0.095 & 3.82/0.109 & \textbf{3.14}/\textbf{0.073} \\
$T_{\mathrm{tot}}$ $\downarrow$ & 2.48 & 2.87 & 3.30 & 1.09 & \textbf{1.07} & 1.26 & 1.19 \\
$SR$ $\uparrow$ & 67.2 & 67.2 & 76.6 & 73.4 & 75.0 & 68.8 & \textbf{81.2} \\
\bottomrule
\end{tabular}
\\[4pt]
{\footnotesize ATE: $^\circ$/m \quad $T_{\mathrm{tot}}$: s \quad $SR$: \%.}
\vspace{-2em}
\end{table*}

We conduct extensive experiments on two popular public datasets: EuRoC MAV~\cite{Burri2016EuRoCmicro} and TUM-VI~\cite{Schubert2018TUMVI}, along with our self-collected dataset. For statistical reliability, following~\cite{Merrill2023FastMonocular}, we divide each sequence into 10-second windows, perform initialization and evaluation in each window, and report the average results. To provide a comprehensive evaluation, we assess the initialization performance using the following metrics:
\begin{enumerate*}[label=(\arabic*)]
\item gravity direction error,
\item linear velocity error, and
\item absolute trajectory error (ATE)~\cite{Zhang2018TutorialQuantitative} over the entire 10-second window
\end{enumerate*}.
Beyond these common metrics, we additionally report the following:
\begin{enumerate}[label=(\arabic*), resume]
\item \textbf{Total time \(T_{\mathrm{tot}}\)}. This metric measures the total temporal span of sensory data consumed by continuous initialization attempts until a successful initialization is achieved.
\item \textbf{Success rate (SR)}. We adopt stricter success criteria than those used in the compared baseline~\cite{Merrill2023FastMonocular}, requiring that:
\begin{enumerate*}[label=(\roman*)]
\item the optimization converges,
\item the state covariance is successfully recovered, and
\item the position absolute trajectory error (ATE) of VINS tracking over the entire window remains below 0.5\,m.
\end{enumerate*}
\item \textbf{Scale error}. This quantifies the relative error of the estimated point cloud scale, where the ground-truth scale is computed as the ratio between the predicted translation displacement and the ground-truth displacement.
\end{enumerate}
These additional metrics provide a practical assessment of the sensory data required for successful initialization and the suitability of the resulting estimates for subsequent tracking.
\vspace{-2pt}
\subsection{Implementation Details}
We compare against the following baselines:
\begin{enumerate}[label=(\arabic*)]
  \item \textbf{Dong-Si}: The default initialization method of OpenVINS~\cite{Dong-Si2012Estimatorinitialization}, combining Eq.~\eqref{eq:linear_feature} and Eq.~\eqref{eq:nonlinear_system_feature}.
  \item \textbf{Dep.(Const.)}: Our re-implementation of~\cite{Zhou2022LearnedMonocular} leveraging learned monocular depth priors while using the Dong-Si method for providing the initial guess.
  \item \textbf{Dep.(AB)}: Our re-implementation of~\cite{Merrill2023FastMonocular}.
  \item \textbf{LC}: Our re-implementation of loosely-coupled method as described in~\cite{Qin2018VINSMonoRobust} by aligning the estimated camera trajectory and preintegrated IMU trajectory, using Eq.~\eqref{eq:nonlinear_system_feature} for refinement.
  \item \textbf{sqrtVINS}: The initialization module of~\cite{Peng2025sqrtmathbfVINS}.
  \item \textbf{DRT}: The rotation-translation-decoupled initialization method~\cite{He2023RotationTranslationDecoupledSolution}.
\end{enumerate}
We denote our proposed method as \textbf{Ours~(SC)}, which uses the proposed scale-constrained nonlinear refinement (see Eq.~\eqref{eq:nonlinear_system_feature_scale}), and \textbf{Ours~(FF)}, which adopts the fully feature-free formulation (see Eq.~\eqref{eq:reproj_feature_free}). Both methods share the same linear initialization (see Eq.~\eqref{eq:linear_system_scale}).

All methods are integrated into the open-source OpenVINS framework~\cite{Geneva2020OpenVINSResearch}, and chi-square tests are applied to reject obviously erroneous solutions. For a fair comparison, the camera pose for \textbf{LC} and the depth for \textbf{Dep.(Const.)} and \textbf{Dep.(AB)} are also predicted by the VGGT model~\cite{Wang2025VGGTVisuala}. The VGGT network runs on an RTX 6000 Ada GPU, with inference taking approximately 0.24 seconds for five frames.
As~\cite{Zhou2022LearnedMonocular} and~\cite{Merrill2023FastMonocular} are not open source, we reproduced these methods according to their respective papers. Note that for \textbf{Dep.(Const.)}, due to the scale normalization strategy employed by VGGT, we are unable to incorporate the shift constraint for the depth factor. Other details follow the original papers.
\subsection{Evaluation on EuRoC Dataset}
We use a setting of 5 keyframes over a 0.5\,s window on the EuRoC MAV dataset~\cite{Burri2016EuRoCmicro}, where the parallax is relatively small and IMU excitation is limited, yielding challenging initialization scenarios. \textbf{LC} is excluded from the comparison, as it fails to converge within such short windows.

Table~\ref{tab:euroc_linear} compares the linear systems in terms of gravity direction and linear velocity errors. The proposed feature-free linear system generally outperforms Dong-Si and Dep.(AB) and achieves the lowest gravity error overall, demonstrating the effectiveness of the proposed feature-free closed-form formulation. By eliminating 3D feature positions from the linear system, all methods except Dong-Si achieve greater robustness to outliers and consistently higher accuracy across sequences. However, the learning-based methods degrade on V1\_03, where motion blur compromises the 3D model predictions. Incorporating RANSAC further yields a modest but consistent improvement, confirming its effectiveness in suppressing outliers.

The overall initialization performance is summarized in Table~\ref{tab:euroc_summary}. Win.\ ATE measures initialization accuracy within the short temporal window, while Traj.\ ATE evaluates the tracking error of subsequent VINS trajectory, serving as an indicator of initialization reliability. Accordingly, $SR$ is determined based on Traj.\ ATE to filter out numerically converged but incorrect results. 
Ours~(FF) achieves the highest success rate, enabling successful initialization and subsequent tracking in over 80\% of windows while requiring less sensory data, with an average $T_{\mathrm{tot}}$ of only 1.2\,s, demonstrating superior efficiency and reliability.
Comparing the two proposed variants, Ours~(SC) achieves the best Win.\ ATE, surpassing all baselines. Ours~(FF), on the other hand, yields higher Win.\ ATE than Ours~(SC), which is expected due to the absence of feature correspondences and thus weaker geometric constraints during optimization. Nevertheless, Ours~(FF) yields the lowest Traj.\ ATE, suggesting that its initialization quality is sufficient to support accurate and stable VINS tracking.

\subsection{Evaluation on TUM-VI Dataset}

\begin{table*}[!htbp]
\centering
\caption{Evaluation of linear system on TUM-VI dataset: gravity error ($^\circ$) $\downarrow$ / velocity error (m/s) $\downarrow$.}
\vspace{-0.5em}
% \small
\label{tab:tumvi_linear}
\begin{tabular*}{\textwidth}{@{\extracolsep{\fill}}lccccccc}
\toprule
Algorithm & room1 & room2 & room3 & room4 & room5 & room6 & Average \\
\midrule
Dong-Si & 8.39 / 0.87 & 4.99 / 0.84 & 3.56 / 0.56 & 5.77 / 0.53 & 12.60 / 1.05 & 4.61 / 0.38 & 6.65 / 0.71 \\
Dep.(AB) & 6.72 / 0.67 & 2.75 / 0.72 & 3.14 / 0.68 & 5.07 / 0.55 & 4.33 / 0.37 & 1.78 / 0.38 & 3.96 / 0.56 \\
Ours w/o RANSAC & 3.35 / 0.32 & 3.23 / 0.49 & 2.92 / 0.55 & 2.60 / 0.33 & 3.84 / 0.26 & 1.95 / 0.41 & 2.99 / \textbf{0.39} \\
Ours & 3.57 / 0.33 & 3.22 / 0.48 & 2.86 / 0.55 & 2.45 / 0.31 & 3.77 / 0.28 & 1.95 / 0.41 & \textbf{2.97 / 0.39} \\
\bottomrule
\end{tabular*}
\vspace{-1em}
\end{table*}

\begin{table}[!htbp]
\centering
\caption{Overall performance on TUM-VI dataset (Setting: 5 keyframes, 0.5\,s window).}
\label{tab:tumvi_summary}
\vspace{-0.5em}
\setlength{\tabcolsep}{2.8pt}
\begin{tabular}{lccccc}
\toprule
 & Dong-Si & Dep.(AB) & Dep.(Const.) & Ours~(SC) & Ours~(FF) \\
\midrule
Win. ATE $\downarrow$ & 1.14/0.024 & 1.11/\textbf{0.021} & 1.14/0.031 & \textbf{1.04}/0.023 & 3.22/0.067 \\
Traj. ATE $\downarrow$ & 1.59/\textbf{0.022} & 1.67/\textbf{0.022} & 2.18/0.045 & \textbf{1.40}/0.032 & 2.10/0.068 \\
$T_{\mathrm{tot}}$ $\downarrow$ & 2.87 & 3.63 & 4.66 & 1.16 & \textbf{1.10} \\
$SR$ $\uparrow$ & 74.7 & 82.7 & 78.7 & 92.0 & \textbf{94.7} \\
\bottomrule
\end{tabular}
\\[4pt]
\parbox{\columnwidth}{\raggedright
ATE: $^\circ$/m \quad
$T_{\mathrm{tot}}$: s \quad
$SR$: \%.}
\vspace{-2em}
\end{table}

We further evaluate on six room sequences of the TUM-VI dataset~\cite{Schubert2018TUMVI}, which contain ground truth from a motion capture system. Since VGGT can only accept undistorted images, we rectify the fisheye image and crop the field of view during the initialization to ensure the prediction quality. We also adopt a setting of 5 keyframes over a 0.5\,s window in this dataset. Compared to EuRoC, TUM-VI provides richer motion excitation, leading to better initialization accuracy in general. As shown in Table~\ref{tab:tumvi_linear}, the advantage of our linear system becomes more pronounced under stronger IMU excitation, achieving approximately 25\% reduction in gravity direction error compared to Dep.(AB). For overall performance in Table~\ref{tab:tumvi_summary}, both variants achieve comparable accuracy with only 1.1\,s data duration, while improving the $SR$ to over 90\%.

\begin{figure}[!htbp]
\centering
\includegraphics[width=0.75\columnwidth]{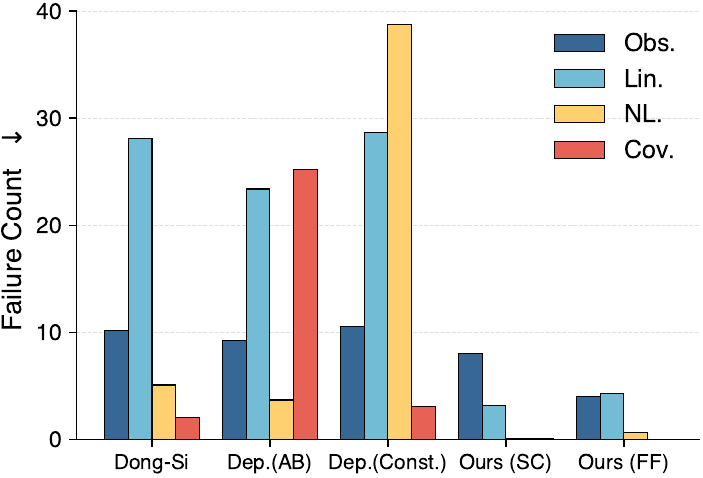}
\caption{Failure source distribution across methods. Obs.: insufficient observations. Lin.: rank-deficient linear system. NL.: divergent optimization. Cov.: covariance recovery failure.}
\label{fig:fail_count}
% \vspace{-1em}
\end{figure}

\begin{figure}[!htbp]
\centering
\includegraphics[width=0.9\columnwidth]{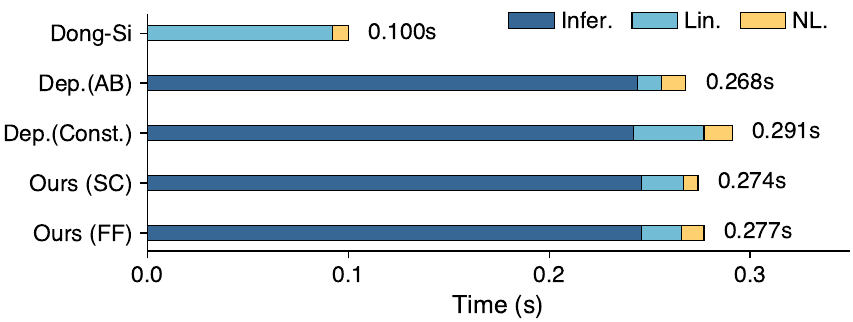}
\caption{Per-attempt initialization runtime breakdown. Infer.: VGGT inference. Lin.: linear system. NL.: nonlinear refinement. Note that the proposed methods need significantly fewer attempts.}
\label{fig:runtime}
\vspace{-0.5em}
\end{figure}

To better understand total data consumption $T_{\mathrm{tot}}$, we analyze the failure sources of each method and categorize them into four types as shown in Fig.~\ref{fig:fail_count}.
%: (1) insufficient observations, (2) rank-deficient linear system, (3) divergent nonlinear optimization, and (4) covariance recovery failure. Fig.~\ref{fig:fail_count} illustrates the distribution of failure sources for different methods. 
All baselines exhibit a large number of failures caused by insufficient observations and rank-deficient linear systems. Although Dep.(AB) eliminates the estimation of feature positions in the linear system, it still relies on feature correspondences implicitly and thus suffers from the same limitations. In addition, Dep.(AB) and Dep.(Const.) show increased failures in covariance recovery and nonlinear optimization respectively, stemming from the underlying feature-based formulation. Owing to a robust closed-form solution and reduced state dimension, Ours~(FF) and Ours~(SC) significantly reduce failures across all categories, thus achieving lowest $T_{\mathrm{tot}}$. 
We also analyze the per-attempt runtime in Fig.~\ref{fig:runtime}. Learning-based methods exhibit similar runtime dominated by VGGT inference, and are slower than Dong-Si which requires no network forward pass. Ours~(FF) takes slightly longer in nonlinear refinement than Ours~(SC), since removing feature correspondences yields a flatter optimization landscape that requires more iterations to converge. However, this comparison reflects only the cost of a single attempt. As shown in Fig.~\ref{fig:fail_count}, the proposed methods require significantly fewer attempts to succeed, thereby achieving the lowest overall computational time for successful initialization.

\subsection{Ablation Study}
To validate individual components of the proposed method, we conduct extensive ablation studies on TUM-VI dataset, as summarized in Table~\ref{tab:tumvi_ablation}. In addition to standard initialization metrics, we report the scale errors estimated from both the linear system and nonlinear refinement stages, reflecting the scale accuracy of the recovered 3D scene. Unless otherwise specified, the default configuration employs the feature-free formulation with 100 sampled points, RANSAC outlier rejection, and region-partitioned scale.

We first analyze the influence of the number of sampled points $K$ in the linear system. As $K$ increases from 10 to 1000, gravity and velocity errors generally decrease, indicating that more points provide better geometric constraints. However, a larger $K$ leads to a higher computational cost and numerical instability, since more low-confidence points may be included. As can be seen when $K = 1000$, it results in longer $T_{\mathrm{tot}}$ and lower $SR$. From an accuracy perspective, Pts.\ 500 achieves the best performance, while we also observe that reasonable accuracy can be attained with only 10 points. In summary, we select $K=100$ for the highest $SR$.

We then evaluate the remaining components. RANSAC slightly improves both $SR$ and accuracy by filtering outliers in the predicted point cloud. Global Scale, which applies a single scale factor to the entire point cloud rather than region-partitioned scales, achieves comparable performance, indicating that VGGT produces consistent scale predictions across the scene. Since LC fails to initialize within the default 0.5\,s window, we additionally report LC (1s) with an extended 1\,s window. Even under this relaxed setting, LC exhibits significantly lower $SR$ and longer $T_{\mathrm{tot}}$, highlighting the advantage of our feature-free formulation. Finally, the comparison between w/ SC and w/o SC validates the effectiveness of scale-constrained nonlinear refinement. Incorporating scale constraints improves both accuracy and $SR$, and yields only 8\% scale error after refinement. An additional ablation on the temporal window length is provided in Tab.~S1 of the supplementary material, showing that longer windows further improve accuracy and success rate. 

\begin{table*}[!htbp]
\centering
% \small
\caption{Ablation study on TUM-VI\@. We evaluate sampled points, RANSAC, regional scale, and scale constraints.}
\label{tab:tumvi_ablation}
\vspace{-0.5em}
\begin{tabular*}{\textwidth}{@{\extracolsep{\fill}}lcccccccc}
\toprule
& \makecell{Grav.\ Err.\\Lin. ($^\circ$) $\downarrow$} & \makecell{Vel.\ Err.\\Lin. (m/s) $\downarrow$} & \makecell{Scale Err.\\Lin.\ (\%) $\downarrow$} & \makecell{Scale Err.\\NL.\ (\%) $\downarrow$} & \makecell{Win.\ ATE\\($^\circ$/m) $\downarrow$} & \makecell{Traj.\ ATE\\($^\circ$/m) $\downarrow$} & \makecell{$T_{\mathrm{tot}}$\\(s) $\downarrow$} & \makecell{SR\\(\%) $\uparrow$} \\
\midrule
Pts.\ 10 & 2.83 & 0.35 & 38.03 & 29.62 & 3.05/0.075 & 1.83/0.058 & 1.10 & 86.7 \\
Pts.\ 20 & 3.08 & 0.40 & 41.83 & 30.08 & 3.35/0.077 & 1.96/0.063 & 1.03 & 89.3 \\
Pts.\ 50 & 3.19 & 0.39 & 43.48 & 28.63 & 3.46/0.071 & 2.15/0.070 & 1.08 & 89.3 \\
\textbf{Pts.\ 100 (Ours)} & \textbf{2.97} & \textbf{0.39} & \textbf{42.48} & \textbf{25.72} & \textbf{3.22/0.067} & \textbf{2.10/0.068} & \textbf{1.10} & \textbf{94.7} \\
Pts.\ 200 & 2.70 & 0.39 & 42.20 & 24.32 & 3.00/0.064 & 2.01/0.061 & 1.26 & 88.0 \\
Pts.\ 500 & 2.46 & 0.37 & 41.10 & 20.27 & 2.69/0.054 & 1.98/0.058 & 1.26 & 86.7 \\
Pts.\ 1000 & 2.47 & 0.40 & 43.50 & 21.38 & 2.67/0.055 & 2.38/0.068 & 1.48 & 80.0 \\
\midrule
w/o RANSAC & 2.99 & 0.39 & 42.13 & 25.12 & 3.22/0.065 & 2.14/0.071 & 1.05 & 88.0 \\
\midrule
Global Scale & 2.89 & 0.40 & 41.13 & 24.87 & 3.17/0.066 & 2.22/0.067 & 1.10 & 89.3 \\
\midrule
LC (1s) & 1.25 & 0.21 & 13.73 & - & 0.81/0.028 & 1.46/0.027 & 4.71 & 40.0 \\
\midrule\midrule
w/ SC & 3.21 & 0.41 & 41.62 & 8.80 & 1.04/0.023 & 1.40/0.032 & 1.16 & 92.0 \\
w/o SC & 4.07 & 0.50 & 47.15 & - & 1.23/0.026 & 1.63/0.035 & 2.88 & 82.7 \\
\bottomrule
\end{tabular*}
\end{table*}

The proposed method is designed to be agnostic to the choice of feed-forward 3D geometry model. To validate this property and to examine the impact of different 3D predictions on initialization performance, we incorporate alternative models into the pipeline. Specifically, we replace the default VGGT~\cite{Wang2025VGGTVisuala} model with Depth Anything 3 (DAv3)~\cite{Lin2025DepthAnything} and $\pi^3$~\cite{Wang2025p3Scalable} while keeping all other components unchanged.

As shown in Table~\ref{tab:model_ablation_linear}, Ours~(DAv3) and Ours~($\pi^3$) consistently achieve lower errors across all sequences than Ours~(VGGT), demonstrating that the proposed initialization framework can directly benefit from improved feed-forward geometric predictions. This confirms that the formulation is not tied to a specific 3D model and naturally scales with the quality of the underlying geometry.
Among the evaluated models, $\pi^3$ yields the best overall performance. We use its latest $\pi^3X$ variant, which supports conditional injection of known camera intrinsics during inference. In contrast, models without this capability rely on internally inferred intrinsics that may deviate from the true camera parameters, leading to additional geometric inconsistencies in the predicted 3D output. Detailed ablations on $\pi^3$ are provided in Tab.~S2 of the supplementary material.

\begin{table*}[!htbp]
\centering
\caption{Ablation of feed-forward 3D model on linear system: gravity error ($^\circ$) $\downarrow$ / velocity error (m/s) $\downarrow$.}
\vspace{-0.5em}

\label{tab:model_ablation_linear}
\begin{tabular*}{\textwidth}{@{\extracolsep{\fill}}lccccccc}
\toprule
Model & room1 & room2 & room3 & room4 & room5 & room6 & Average \\
\midrule
Ours~(VGGT) & 3.57 / 0.33 & 3.22 / 0.48 & 2.86 / 0.55 & 2.45 / 0.31 & 3.77 / 0.28 & 1.95 / 0.41 & 2.97 / 0.39 \\
Ours~(DAv3) & 1.84 / 0.21 & 2.32 / 0.59 & 2.21 / 0.55 & 2.29 / 0.31 & 1.53 / 0.15 & 1.36 / 0.27 & 1.92 / 0.35 \\
Ours~($\pi^3$) & 2.08 / 0.25 & 1.57 / 0.34 & 2.20 / 0.43 & 1.41 / 0.15 & 2.54 / 0.22 & 1.02 / 0.24 & \textbf{1.80 / 0.27} \\
\bottomrule
\end{tabular*}
\end{table*}

\subsection{Evaluation on Self-Collected Dataset}
\begin{figure}[!htbp]
\centering
\subfloat[Indoor-Structured]{\includegraphics[width=0.32\columnwidth]{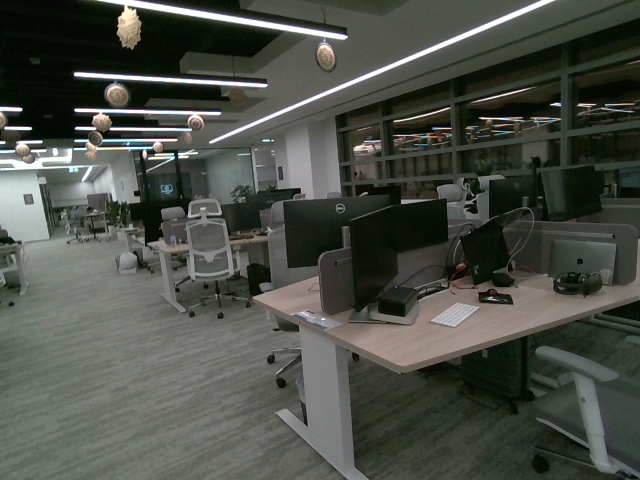}}
\hfill
\subfloat[Indoor-Degraded]{\includegraphics[width=0.32\columnwidth]{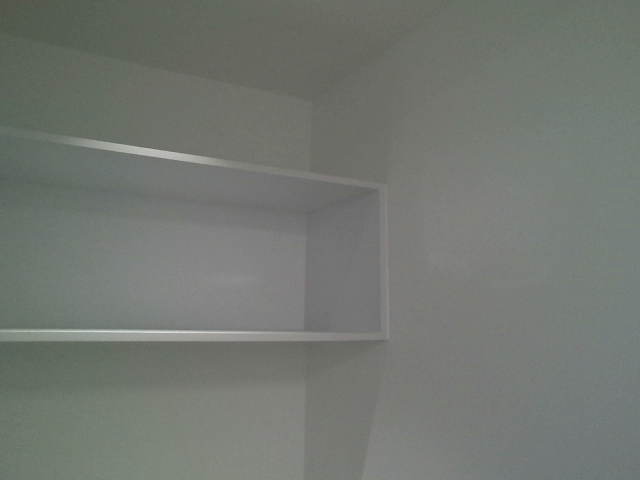}}
\hfill
\subfloat[Outdoor]{\includegraphics[width=0.32\columnwidth]{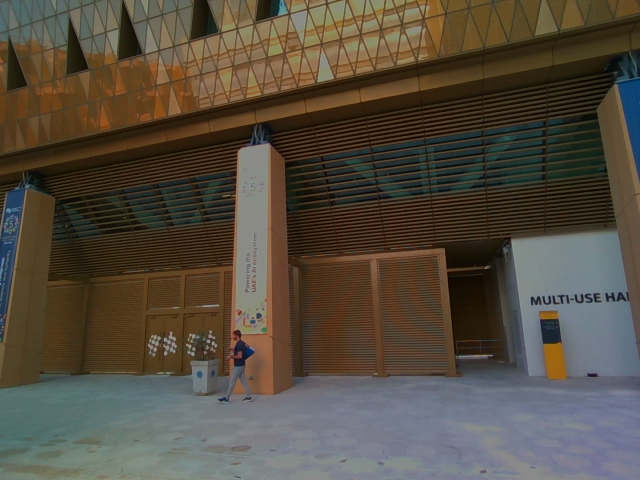}}
\caption{Example scenes from our self-collected dataset: (a) indoor office with complex depth, (b) visually degraded indoor environment, and (c) outdoor.}
\label{fig:dataset_scenes}
% \vspace{-1em}
\end{figure}

To validate the generalization and applicability of the feature-free formulation, we collect a dataset that contains various challenging real-world scenarios, as shown in Fig.~\ref{fig:dataset_scenes}.
The data are recorded by a RealSense D455 camera, which provides RGB images at 30\,Hz and consumer-grade IMU measurements at 200\,Hz. Each sequence is approximately 25 seconds long, with figure-eight motion patterns designed specifically for initialization validation. Since no ground truth is available, we qualitatively evaluate initialization success by examining whether the VINS pose tracking severely drifts after initialization.
Due to the challenging nature of the scenarios, all methods use a 2\,s initialization window. Table~\ref{tab:self_collected} shows qualitative results.

\begin{table*}[t]
\centering
\setlength{\tabcolsep}{14pt}
\caption{Success rate evaluation on self-collected dataset.}
\vspace{-0.5em}
\label{tab:self_collected}
\begin{tabular}{llccccccc}
\toprule
Category & Seq. & Ours~(FF) & Dep.(Const.) & Dep.(AB) & Dong-Si & LC & sqrtVINS & DRT \\
\midrule
\multirow{2}{*}{\makecell{Indoor-\\Structured}}
 & office\_01 & \cmark & \cmark & \cmark & \cmark & \cmark & \cmark & \cmark \\
 & office\_02 & \cmark & \cmark & \cmark & \cmark & \cmark & \cmark & \cmark \\
\midrule
\multirow{4}{*}{\makecell{Indoor-\\Degraded}}
 & printer\_01 & \cmark & \xmark & \cmark & \xmark & \xmark & \xmark & \xmark \\
 & printer\_02 & \cmark & \xmark & \xmark & \xmark & \xmark & \xmark & \xmark \\
 & lift\_01 & \cmark & \xmark & \xmark & \xmark & \xmark & \cmark & \cmark \\
 & lift\_02 & \cmark & \xmark & \xmark & \xmark & \xmark & \xmark & \xmark\\
\midrule
\multirow{4}{*}{Outdoor}
 & outdoor\_01 & \cmark & \xmark & \xmark & \cmark & \xmark & \cmark & \xmark\\
 & outdoor\_02 & \cmark & \cmark & \xmark & \xmark & \xmark & \xmark & \xmark\\
 & outdoor\_03 & \cmark & \cmark & \cmark & \xmark & \cmark & \cmark & \cmark\\
 & outdoor\_04 & \xmark & \xmark & \xmark & \xmark & \xmark & \xmark & \xmark\\
\midrule
\multicolumn{2}{c}{$SR$ (\%) $\uparrow$} & \textbf{90} & 40 & 40 & 30 & 30 & 50 & 40 \\
\bottomrule
\end{tabular}
\\[2pt]
{\footnotesize \cmark: successful initialization without obvious drift. \xmark: failure or severe drift.}
\vspace{-0.5em}
\end{table*}

\begin{figure}[!htbp]
\centering
\includegraphics[width=1\columnwidth]{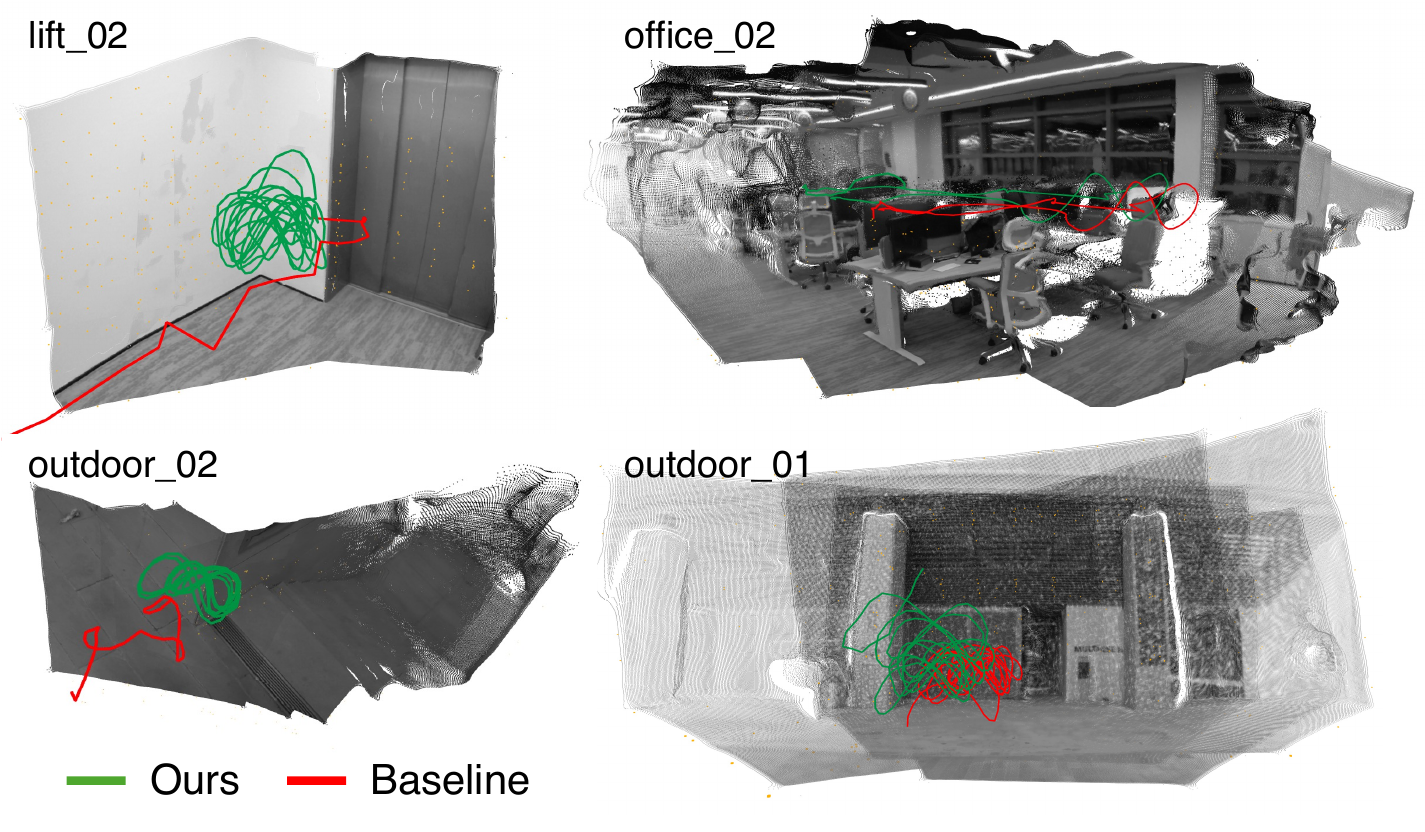}
\caption{Trajectory comparison on representative sequences. Green: Ours. Red: Baseline (only one shown for clarity). The baseline exhibits significant drift in degraded scenarios, while Ours~(FF) maintains stable tracking across all categories.}
\label{fig:qualitative}
\vspace{-1em}
\end{figure} 
In Indoor-Structured scenarios, all learning-based methods succeed, demonstrating strong geometric inference capability of the 3D foundation model.
In Indoor-Degraded scenarios, only 50--60 features are detected per frame on average. Under such conditions, all baselines often diverge severely or even fail to satisfy initialization requirements, whereas our feature-free formulation succeeds reliably, as shown in Fig.~\ref{fig:qualitative}. However, we find that Ours~(FF) might also fail in extremely degraded environments (e.g., blank white walls), as the input images lack sufficient geometric cues for the 3D model.
In Outdoor scenarios, all methods fail on `outdoor\_04' sequence due to the open-sky scene, where the 3D model produces unreliable predictions. Further dataset samples and additional qualitative trajectory comparisons across all sequences are provided in Fig.~S2 and Fig.~S3 of the supplementary material. 

\section{Conclusion and Future Work}
\label{sec:conclusion}
We present a feature-free monocular visual-inertial initialization framework that leverages point clouds predicted by feed-forward 3D models. Using the predicted scene geometry, we derive an efficient closed-form linear solver and a compact nonlinear refinement formulation. 
Extensive experiments on public datasets demonstrate that our method achieves success rates exceeding 90\% while requiring only about 1.2\,s of sensory data for initialization.
The low scale error further indicates that the metric scale of the 3D model can be reliably recovered. Results on a self-collected dataset further validate the generalization across various scenarios, with robust performance particularly in visually degraded environments. 

Our method still has limitations. While the feature-free formulation achieves higher efficiency and success rates than its scale-constrained counterpart, it can yield slightly lower pose accuracy; in practice, the two formulations can be selected according to application requirements. A more fundamental limitation lies in the generalization of current feed-forward 3D foundation models: in open-sky scenes with sparse or ambiguous visual cues, the predicted geometry becomes unreliable and degrades initialization, which is a challenge shared by all learning-based baselines evaluated here. Since our formulation is model-agnostic, future improvements in foundation models will directly translate into better initialization. Developing a fully differentiable, feed-forward VINS is another promising direction toward further robustness.

\bibliographystyle{IEEEtranN}
\bibliography{references}

\clearpage

% ===== Supplementary Material =====
\section*{Supplementary Material}

% S-prefix for figures/tables/equations in the supplementary, matching SupplementaryMaterial.tex
\renewcommand{\thefigure}{S\arabic{figure}}
\renewcommand{\thetable}{S\arabic{table}}
\renewcommand{\theequation}{S\arabic{equation}}
\setcounter{figure}{0}
\setcounter{table}{0}
\setcounter{equation}{0}

\section{Method Details}
\label{sec:method_details}
In this section, we provide additional algorithmic details of the proposed feature-free method. 
\subsection{Rank Analysis of Feature-Free Linear System}
This section analyzes the rank properties of the proposed feature-free linear system to determine the minimal conditions required for state estimation. In our formulation, the state has dimension $1+3+3=7$.
For completeness and ease of reference, we restate the key equations used in the main paper:
\begin{align}
\bar{\mathbf{A}}\mathbf{x} = \bar{\mathbf{b}},\, \mathbf{x} =
\begin{bmatrix}
s & {}^{I_0}\mathbf{v}_{I_0}^\top & {}^{I_0}\mathbf{g}^\top
\end{bmatrix}^\top
\label{eq:linear_state_vector_scale_appendix}
\end{align}
\begin{align}
\bar{\mathbf{A}}_{i,k} &= \mathbf{H}_{i,k}
\begin{bmatrix}
{}^{I}_{C}\mathbf{R}\, {}^{C_0}\bar{\mathbf{p}}_{o_{1,k}} & -\Delta t\,\mathbf{I}_3 & \frac{1}{2}\Delta t^2\,\mathbf{I}_3
\end{bmatrix} \notag \\
\bar{\mathbf{b}}_{i,k} &= \mathbf{H}_{i,k}\left({}^{I_0}\Delta \mathbf{p}_{I_i} - {}^{I_i}_{I_0}\mathbf{R}\,{}^{I}_{C}\mathbf{R}\,{}^{C}\mathbf{p}_{I} + {}^{I}_{C}\mathbf{R}\,{}^{C}\mathbf{p}_{I}\right) \label{eq:linear_system_scale_appendix}
\end{align}
\textbf{Case 1: Two images} \\
Given two image frames $t_0$ and $t_1$, we initialize the VIO system by stacking the $k$ linear constraints in Eq.~\eqref{eq:linear_system_scale_appendix}, resulting in the following linear system. 
\begin{align}
\resizebox{0.6\columnwidth}{!}{$
\begin{aligned}
\mathbf{\bar{A}}
&=
\operatorname{diag}\!\bigl(
\mathbf{\mathbf{H}}_{1,1},\mathbf{\mathbf{H}}_{1,2},\dots,\mathbf{\mathbf{H}}_{i,k}
\bigr)\;\tilde{\mathbf{A}}, \\[6pt]
\tilde{\mathbf{A}}
&=
\begin{bmatrix}
{}^{I}_{C}\mathbf{R}\, {}^{C_0}\bar{\mathbf{p}}_{o_{1,1}} &
-\Delta t_1 \mathbf{I} &
\tfrac{1}{2}\Delta t_1^2\mathbf{I} \\[8pt]
{}^{I}_{C}\mathbf{R}\, {}^{C_0}\bar{\mathbf{p}}_{o_{1,2}} &
-\Delta t_1 \mathbf{I} &
\tfrac{1}{2}\Delta t_1^2\mathbf{I} \\[4pt]
\vdots & \vdots & \vdots \\[4pt]
{}^{I}_{C}\mathbf{R}\, {}^{C_0}\bar{\mathbf{p}}_{o_{1,k}} &
-\Delta t_1 \mathbf{I} &
\tfrac{1}{2}\Delta t_1^2 \mathbf{I}
\end{bmatrix}
\end{aligned}
$}
\end{align}
We then analyze the rank of the $\tilde{\mathbf{A}}$ matrix using elementary transformations. Specifically, we add $\frac{1}{2}\Delta t$ times the second column to the third column, which eliminates the third column:
\begin{align}
\resizebox{0.5\columnwidth}{!}{$
\begin{aligned}
& c_3 \;\to\; \tfrac{1}{2}\,\Delta t\, c_2 + c_3 \\[6pt]
&\Rightarrow\;
\begin{bmatrix}
    {}^{I}_{C}\mathbf{R}\, {}^{C_0}\bar{\mathbf{p}}_{o_{1,1}}
    & -\Delta t_1 \mathbf{I}
    & \mathbf{0} \\[6pt]
    {}^{I}_{C}\mathbf{R}\, {}^{C_0}\bar{\mathbf{p}}_{o_{1,2}}
    & -\Delta t_1 \mathbf{I}
    & \mathbf{0} \\[6pt]
    \vdots & \vdots & \vdots \\[6pt]
    {}^{I}_{C}\mathbf{R}\, {}^{C_0}\bar{\mathbf{p}}_{o_{1,k}}
    & -\Delta t_1 \mathbf{I}
    & \mathbf{0}
\end{bmatrix}
\end{aligned}
$}
\end{align}
The transformation result is as follows:
$\operatorname{rank}(\mathbf{\bar{A}}) = 4 < 7$, indicating that with two images, this system is rank-deficient, regardless of the number of points. This can be intuitively understood as the coupling relation between velocity ${}^{I_0}\mathbf{v}_{I_0}$ and gravity ${}^{I_0}\mathbf{g}$.
Thus we claim that, despite the extremely reduced state vector, the observability boundary of visual-inertial initialization remains unchanged, which requires at least three frames. 
\\
\textbf{Case 2: Three images} \\
Let's introduce the third image frame $t_2$, the state vector remains the same, the corresponding $\tilde{\mathbf{A}}$ is
\begin{align}
\resizebox{0.6\columnwidth}{!}{$
\begin{aligned}
\tilde{\mathbf{A}} = 
\begin{bmatrix}
    {}^{I}_{C}\mathbf{R}\, {}^{C_0}\bar{\mathbf{p}}_{o_{1,1}}
    & -\Delta t_1 \mathbf{I}
    & \tfrac{1}{2} \Delta t_1^2 \mathbf{I} \\[6pt]
    {}^{I}_{C}\mathbf{R}\, {}^{C_0}\bar{\mathbf{p}}_{o_{1,2}}
    & -\Delta t_1 \mathbf{I}
    & \tfrac{1}{2} \Delta t_1^2 \mathbf{I} \\[6pt]
    \vdots & \vdots & \vdots \\[6pt]
    {}^{I}_{C}\mathbf{R}\, {}^{C_0}\bar{\mathbf{p}}_{o_{1,k}}
    & -\Delta t_1 \mathbf{I}
    & \tfrac{1}{2} \Delta t_1^2 \mathbf{I} \\[6pt]
    %------Frame j-----%
    {}^{I}_{C}\mathbf{R}\, {}^{C_0}\bar{\mathbf{p}}_{o_{2,1}}
    & -\Delta t_2 \mathbf{I}
    & \tfrac{1}{2} \Delta t_2^2 \mathbf{I} \\[6pt]
    {}^{I}_{C}\mathbf{R}\, {}^{C_0}\bar{\mathbf{p}}_{o_{2,2}}
    & -\Delta t_2 \mathbf{I}
    & \tfrac{1}{2} \Delta t_2^2 \mathbf{I} \\[6pt]
    \vdots & \vdots & \vdots \\[6pt]
    {}^{I}_{C}\mathbf{R}\, {}^{C_0}\bar{\mathbf{p}}_{o_{2,k}}
    & -\Delta t_2 \mathbf{I}
    & \tfrac{1}{2} \Delta t_2^2 \mathbf{I}
    \end{bmatrix} 
\end{aligned}
$}
\end{align}
\textbf{Note.} The index $k$ denotes the $k$-th point sampled from each image frame and does not imply any inter-frame correspondence. 

We then apply elementary row transformations by subtracting rows from frame $t_1$ from the corresponding rows of frame $t_2$, yielding difference terms:
\begin{align} 
\resizebox{0.8\columnwidth}{!}{$
\begin{aligned}
& r_{k+l} \;\to\;  r_{k+l} - r_l  \quad l=1,\dots,k \\[6pt]
&\Rightarrow\;
\begin{bmatrix}
    {}^{I}_{C}\mathbf{R}\, {}^{C_0}\bar{\mathbf{p}}_{o_{1,1}}
    & -\Delta t_1 \mathbf{I}
    & \tfrac{1}{2} \Delta t_1^2 \mathbf{I} \\[6pt]
    {}^{I}_{C}\mathbf{R}\, {}^{C_0}\bar{\mathbf{p}}_{o_{1,2}}
    & -\Delta t_1 \mathbf{I}
    & \tfrac{1}{2} \Delta t_1^2 \mathbf{I} \\[6pt]
    \vdots & \vdots & \vdots \\[6pt]
    {}^{I}_{C}\mathbf{R}\, {}^{C_0}\bar{\mathbf{p}}_{o_{1,k}}
    & -\Delta t_1 \mathbf{I}
    & \tfrac{1}{2} \Delta t_1^2 \mathbf{I} \\[6pt]
    %------Frame j-----%
    {}^{I}_{C}\mathbf{R}\, {}^{C_0}\Delta\bar{\mathbf{p}}_{12,1}
    & (\Delta t_1-\Delta t_2) \mathbf{I}
    & \tfrac{1}{2} (\Delta t_2^2 - \Delta t_1^2) \mathbf{I} \\[6pt]
    {}^{I}_{C}\mathbf{R}\, {}^{C_0}\Delta\bar{\mathbf{p}}_{12,2}
    & (\Delta t_1-\Delta t_2) \mathbf{I}
    & \tfrac{1}{2} (\Delta t_2^2 - \Delta t_1^2) \mathbf{I} \\[6pt]
    \vdots & \vdots & \vdots \\[6pt]
    {}^{I}_{C}\mathbf{R}\, {}^{C_0}\Delta\bar{\mathbf{p}}_{12,k}
    & (\Delta t_1-\Delta t_2) \mathbf{I}
    & \tfrac{1}{2} (\Delta t_2^2 - \Delta t_1^2) \mathbf{I}
\end{bmatrix}
\end{aligned}
$}
\end{align}
where ${}^{C_0}\Delta\bar{\mathbf{p}}_{12,k} = {}^{C_0}\bar{\mathbf{p}}_{o_{2,k}} - {}^{C_0}\bar{\mathbf{p}}_{o_{1,k}}$ denotes the inter-frame point difference. We further subtract the first row of frame $t_2$ from all other rows of frame $t_2$, eliminating the velocity and gravity columns:
\begin{align}
\resizebox{0.95\columnwidth}{!}{$
\begin{aligned}
  & r_{k+l} \;\to\; r_{k+l} - r_{k+1} \quad l=2,\dots,k \\[6pt]
  &\Rightarrow\;
\begin{bmatrix}
    {}^{I}_{C}\mathbf{R}\, {}^{C_0}\bar{\mathbf{p}}_{o_{1,1}}
    & -\Delta t_1 \mathbf{I}
    & \tfrac{1}{2} \Delta t_1^2 \mathbf{I} \\[6pt]
    {}^{I}_{C}\mathbf{R}\, {}^{C_0}\bar{\mathbf{p}}_{o_{1,2}}
    & -\Delta t_1 \mathbf{I}
    & \tfrac{1}{2} \Delta t_1^2 \mathbf{I} \\[6pt]
    \vdots & \vdots & \vdots \\[6pt]
    {}^{I}_{C}\mathbf{R}\, {}^{C_0}\bar{\mathbf{p}}_{o_{1,k}}
    & -\Delta t_1 \mathbf{I}
    & \tfrac{1}{2} \Delta t_1^2 \mathbf{I} \\[6pt]
    %------Frame j-----%
    {}^{I}_{C}\mathbf{R}\, {}^{C_0}\Delta\bar{\mathbf{p}}_{12,1}
    & (\Delta t_1-\Delta t_2) \mathbf{I}
    & \tfrac{1}{2} (\Delta t_2^2 - \Delta t_1^2) \mathbf{I} \\[6pt]
    {}^{I}_{C}\mathbf{R}\, ({}^{C_0}\Delta\bar{\mathbf{p}}_{12,2} - {}^{C_0}\Delta\bar{\mathbf{p}}_{12,1})
    & \mathbf{0}
    & \mathbf{0} \\[6pt]
    \vdots & \vdots & \vdots \\[6pt]
    {}^{I}_{C}\mathbf{R}\, ({}^{C_0}\Delta\bar{\mathbf{p}}_{12,k} - {}^{C_0}\Delta\bar{\mathbf{p}}_{12,1})
    & \mathbf{0}
    & \mathbf{0}
\end{bmatrix}
\end{aligned}
$}
\end{align}
where ${}^{I}_{C}\mathbf{R}{}^{C_0}\Delta\bar{\mathbf{p}}_{12,k} = {}^{I}_{C}\mathbf{R}\,({}^{C_0}\bar{\mathbf{p}}_{o_{2,k}} - {}^{C_0}\bar{\mathbf{p}}_{o_{1,k}})$ denotes the difference between visual bearing vectors. In contrast to feature-based formulations, where corresponding points yield identical bearing vectors and this term vanishes, the difference term here is generally nonzero since no correspondence is used. Hence our feature-free formulation provides better observability than feature-based formulations.
From the transformed matrix, we observe that with $N \geq 3$ frames and sufficiently diverse non-degenerate point measurements, $\bar{\mathbf{A}}$ can reach full column rank.

This settles the minimal case of proposed feature-free linear system that three images with two points are necessary for our feature-free formulation.

\subsection{RANSAC}
Due to the fixed state dimension of our feature-free formulation, RANSAC can be directly integrated into the solving process of linear system to reject outliers in the predicted point cloud. Although theoretically the minimal condition requires only 2 measurements from 3 frames, we empirically found that relatively more points can bring better numerical stability. To this end, we set the minimal subset $\mathcal{S}$ to contain 10 points from every frame in our implementation. 
The overall procedure is summarized in Algorithm~\ref{alg:ransac_linear_init}.
\begin{algorithm}[!htbp]
\caption{RANSAC for Linear System}
\label{alg:ransac_linear_init}
\KwIn{Linear system rows $\{\bar{\mathbf{A}}_{i,m}, \bar{\mathbf{b}}_{i,m}\}$ from Eq.~\eqref{eq:linear_system_scale_appendix}; iterations $Q$; threshold $\tau$}
\KwOut{Robust state estimate $\mathbf{x}^{*}$ from Eq.~\eqref{eq:linear_state_vector_scale_appendix}}
$e^{*} \leftarrow \infty$\;
\For{$q=1$ \KwTo $Q$}{
  $\mathcal{S} \leftarrow$ randomly select minimal subset\;
  $\mathbf{x} \leftarrow \textsc{Solve}(\bar{\mathbf{A}}_{\mathcal{S}}, \bar{\mathbf{b}}_{\mathcal{S}})$\;
  \tcp{Find inliers}
  \ForEach{$(i,m) \notin \mathcal{S}$}{
    \If{$\| \bar{\mathbf{A}}_{i,m} \mathbf{x} - \bar{\mathbf{b}}_{i,m} \| < \tau$}{
      $\mathcal{S} \leftarrow \mathcal{S} \cup \{(i,m)\}$\;
    }
  }
  \tcp{Refit and update}
  $\mathbf{x} \leftarrow \textsc{Solve}(\bar{\mathbf{A}}_{\mathcal{S}}, \bar{\mathbf{b}}_{\mathcal{S}})$\;
  \If{$\| \bar{\mathbf{A}}_{\mathcal{S}} \mathbf{x} - \bar{\mathbf{b}}_{\mathcal{S}} \| < e^{*}$}{
    $e^{*} \leftarrow \| \bar{\mathbf{A}}_{\mathcal{S}} \mathbf{x} - \bar{\mathbf{b}}_{\mathcal{S}} \|$,\; $\mathbf{x}^{*} \leftarrow \mathbf{x}$\;
  }
}
\Return{$\mathbf{x}^{*}$}\;
\end{algorithm}

The inlier set $\mathcal{S}^{*}$ is directly reused in the subsequent nonlinear refinement, providing a consistent set of reliable measurements across both stages.

\subsection{Jacobians for Feature-Free Nonlinear Optimization}
\label{supp:jacobians}
In the feature-free formulation, the reprojection residual is parameterized by the scale factor $s$ instead of explicit 3D point coordinates. The residual is defined as:
\begin{align}
\mathbf{r}_{\mathcal{C}}({}^{C_0}\bar{\mathbf{p}}_{o_{1,k}}, \mathbf{x}_{I_i}, s) &= \mathbf{z}_{i,k}^{\mathcal{C}} - \mathbf{\hat{z}}_{i,k}^{\mathcal{C}}
\end{align}
where the predicted measurement $\mathbf{\hat{z}}_{i,k}^{\mathcal{C}}$ is progressively expanded by coordinate frame transformations:
\begin{align}
\mathbf{\hat{z}}_{i,k}^{\mathcal{C}} &= \pi({}^{C_i}{\mathbf{p}}_{o_{i,k}}) \notag \\ 
% &= \pi\big({}^{C}_{I}\mathbf{T} \, {}^{I_i}_{G}\mathbf{T} \, {}^{G}_{I_0}\mathbf{T} \, {}^{I}_{C}\mathbf{T} \, (s \, {}^{C_0}\bar{\mathbf{p}}_{o_{1,k}})\big) \\ \notag
&= \pi \big( {}^{C}_{I}\mathbf{R}{}^{I}\mathbf{p}_{o_{i,k}} + {}^{C}\mathbf{p}_{I} \big) \notag \\ 
&= \pi \big( {}^{C}_{I}\mathbf{R}{}^{I_i}_{G}\mathbf{R}\left({}^{G}\mathbf{p}_{o_{i,k}}-{}^{G}\mathbf{p}_{I_i}\right) + {}^{C}\mathbf{p}_{I} \big) 
\label{eq:reproj_feature_free_detail}
\end{align}
Here, with ${}^{G}_{I_0}\mathbf{R}$ known from preintegration, the 3D point in the global frame is parameterized solely by the scale $s$:
\begin{align}
{}^{G}\mathbf{p}_{o_{i,k}} = {}^{G}_{I_0} \mathbf{R} \left( {}^{I}_{C}\mathbf{R} \, (s \, {}^{C_0}\bar{\mathbf{p}}_{o_{1,k}}) + {}^{I}\mathbf{p}_{C} \right)
\end{align}
Moreover, to prevent degeneracy as $s$ approaches 0, we parameterize the scale as:
\begin{align}
    s = \varepsilon + \log\left(\exp(\tilde{s}) + 1\right), \quad \varepsilon = 10^{-5}
\end{align}
\textbf{Jacobian w.r.t.\ scale:} \\
The Jacobian can be computed via the chain rule according to Eq.~\eqref{eq:reproj_feature_free_detail}.
\begin{align}
    \frac{\partial \mathbf{r}}{\partial \tilde{s}} 
    = & - \frac{\partial \mathbf{r}}{\partial {}^{C}\mathbf{p}_{o_{i,k}}} \, \frac{\partial {}^{C}\mathbf{p}_{o_{i,k}}}{\partial {}^{G}\mathbf{p}_{o_{i,k}}} \,\frac{\partial {}^{G}\mathbf{p}_{o_{i,k}}}{\partial s} \, \frac{\partial s}{\partial \tilde{s}} \notag \\
    = & - \mathbf{J}_\pi {}^{C}_{I}\mathbf{R} \, {}^{I_i}_{G}\mathbf{R}{}^{G}_{I_0}\mathbf{R} \, {}^{I}_{C}\mathbf{R} \, {}^{C_0}\bar{\mathbf{p}}_{o_{i,k}} \, \frac{e^{\tilde{s}}}{1+e^{\tilde{s}}}
\end{align}
where $\mathbf{J}_\pi$ denotes standard projection jacobian.
\begin{align}
\mathbf{J}_\pi = \frac{\partial \mathbf{r}}{\partial {}^{C}\mathbf{p}_{o_{i,k}}} =
\begin{bmatrix}
1/z & 0 & -x/z^2 \\
0 & 1/z & -y/z^2
\end{bmatrix}
\end{align}

\section{Additional Experimental Results}
\begin{figure*}[!htbp]
\centering
\includegraphics[width=\textwidth]{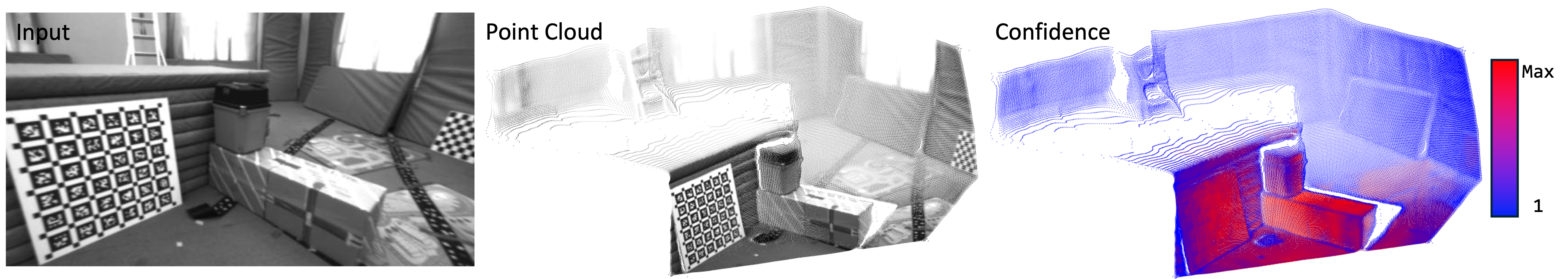}
\caption{Inference results. From left to right: input image, predicted point cloud, and corresponding confidence score (brighter indicates higher confidence).}
\label{fig:vggtinfer}
\end{figure*}

In this section, we present additional experimental results to further analyze the proposed method. We first visualize the feed-forward 3D model inference in Fig.~\ref{fig:vggtinfer}, showing the predicted point cloud and confidence $\{ {}^{C_0}\bar{\mathbf{p}}_{o_{i,k}}, c_{i,k} \}$ used for initialization. We then describe the evaluation metrics and report ablation studies on the temporal window size and the choice of feed-forward 3D model, providing insights into practical configurations and generalization behavior.

\label{sec:additional_results}
\subsection{Evaluation Metric Details} 
\paragraph{Gravity Direction Error} We evaluate the accuracy of gravity estimation using the angular error between the estimated and ground-truth gravity directions.
Assuming the gravity direction is aligned with the global frame $z$-axis, i.e., $\mathbf{e}_z = [0,0,1]^\top$, the gravity vectors expressed in the body frame are obtained as:
\begin{equation}
\hat{\mathbf{g}} = \mathbf{R}_{\text{est}}^\top \mathbf{e}_z, \qquad
\mathbf{g} = \mathbf{R}_{\text{gt}}^\top \mathbf{e}_z,
\end{equation}
where $\mathbf{R}_{\text{est}}$ and $\mathbf{R}_{\text{gt}}$ denote the estimated and ground-truth rotations, respectively.
The gravity direction error is then defined as the angular difference between the two normalized vectors:
\begin{equation}
e_g
=
\arccos\!\left(
\frac{
\hat{\mathbf{g}}^\top \mathbf{g}
}{
\|\hat{\mathbf{g}}\| \, \|\mathbf{g}\|
}
\right),
\end{equation}
which measures the misalignment of gravity directions and is invariant to the gravity magnitude.

\paragraph{Scale Error} As ground-truth point clouds are unavailable, we evaluate scale accuracy via the estimated camera motion. Because the predicted point cloud and the predicted camera poses are expressed in a shared metric scale, pose scale errors directly indicate point cloud scale errors. We compute the relative scale error as the ratio between the predicted translation displacement and the corresponding ground-truth displacement.
\subsection{Ablation of Window Size}
\label{supp:window_ablation}
In the main experiments, we adopt an extreme initialization setting with a short temporal window of 0.5\,s to demonstrate the efficiency of the proposed method. In this section, we conduct an ablation study on the window size to evaluate initialization performance under more relaxed conditions. By increasing the temporal window length, we analyze how additional motion excitation affects robustness and accuracy, thereby providing more practical guidance for real-world deployment.
\begin{table*}[!htbp]
\centering
\setlength{\tabcolsep}{8pt}
\caption{Ablation study of window size.}
\label{tab:window_ablation}
\begin{tabular}{llccccccccc}
\toprule
& \makecell{Win.\\(s)} & \makecell{Grav.\\($^\circ$) $\downarrow$} & \makecell{Vel.\\(m/s) $\downarrow$} & \makecell{Scale\\Lin.\ (\%) $\downarrow$} & \makecell{Scale\\NL.\ (\%) $\downarrow$} & \makecell{Win.\ ATE\\($^\circ$/m) $\downarrow$} & \makecell{Traj.\ ATE\\($^\circ$/m) $\downarrow$} & \makecell{$T_{\mathrm{tot}}$\\(s) $\downarrow$} & \makecell{SR\\(\%) $\uparrow$} \\
\midrule
\multirow{4}{*}{Ours~(SC)}
& 0.5 & 3.21 & 0.41 & 41.62 & 8.80 & 1.04/0.023 & 1.40/0.032 & 1.16 & 92.0 \\
& 1.0 & 1.66 & 0.30 & 27.63 & 9.03 & 1.34/0.045 & 1.09/0.029 & 1.69 & 93.3 \\
& 1.5 & 1.03 & 0.20 & 20.18 & 9.93 & 1.30/0.062 & 1.27/0.027 & 2.62 & 94.7 \\
& 2.0 & 0.68 & 0.20 & 18.77 & 10.07 & 1.13/0.089 & 1.22/0.037 & 4.41 & 97.3 \\
\midrule
\multirow{4}{*}{Ours~(FF)}
& 0.5 & 2.97 & 0.39 & 42.48 & 25.72 & 3.22/0.067 & 2.10/0.068 & 1.10 & 94.7 \\
& 1.0 & 1.59 & 0.29 & 23.38 & 11.78 & 1.96/0.071 & 1.36/0.036 & 1.94 & 98.7 \\
& 1.5 & 1.29 & 0.25 & 18.98 & 11.92 & 1.79/0.102 & 1.20/0.027 & 3.10 & 100.0 \\
& 2.0 & 1.09 & 0.29 & 21.83 & 13.35 & 1.69/0.152 & 1.21/0.036 & 4.13 & 100.0 \\
\bottomrule
\end{tabular}
\end{table*}

As shown in Table~\ref{tab:window_ablation}, extending the initialization window consistently improves accuracy for both variants, while also increasing the overall computation time due to the larger number of frames involved. This reflects the inherent trade-off between efficiency and robustness in initialization. With longer temporal windows, the success rate increases for both methods, and \textit{Ours~(FF)} achieves a 100\% success rate when the window length exceeds 1\,s, indicating strong robustness under more relaxed initialization conditions. 

The gravity direction error, velocity error, and scale error of the linear system decrease substantially as the window length increases, benefiting from richer IMU excitation. In particular, the gravity direction error is reduced to approximately 1$^\circ$, highlighting the high accuracy of the proposed feature-free closed-form solution.

\subsection{Detailed Ablation with $\pi^3$ Model}
\label{supp:pi3_ablation}
Building on the model comparison reported in the main paper, we perform detailed ablation studies using $\pi^3$ as the underlying 3D model (Table~\ref{tab:pi3_ablation}), following the same experimental settings with 100 sampled points as the default configuration. The observed trends are consistent with those obtained using VGGT: increasing the number of sampled points improves the accuracy of the linear system, and RANSAC contributes to higher success rates. Overall, initialization with $\pi^3$ achieves higher accuracy with slightly reduced total data duration and an improved success rate. 
These results demonstrate that the proposed framework effectively leverages stronger feed-forward 3D models while maintaining efficient behavior, further supporting its model-agnostic design.

\begin{figure*}[!htbp]
\centering
\begin{tabular}{cccc}
\includegraphics[width=0.23\textwidth]{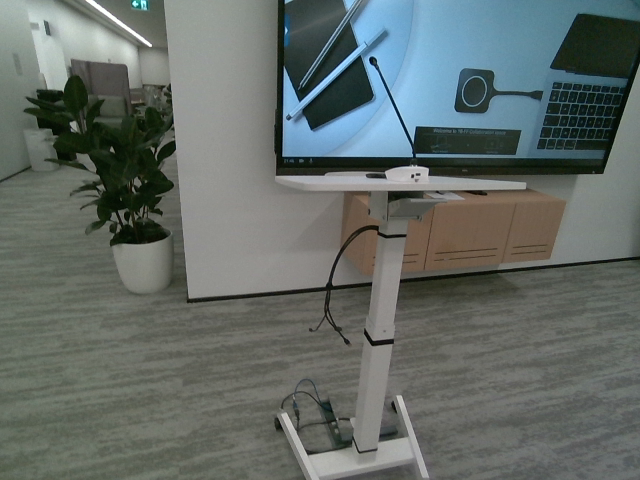} &
\includegraphics[width=0.23\textwidth]{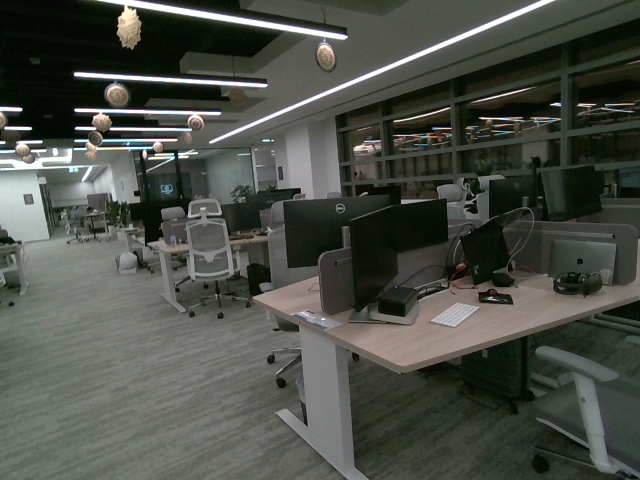} &
\includegraphics[width=0.23\textwidth]{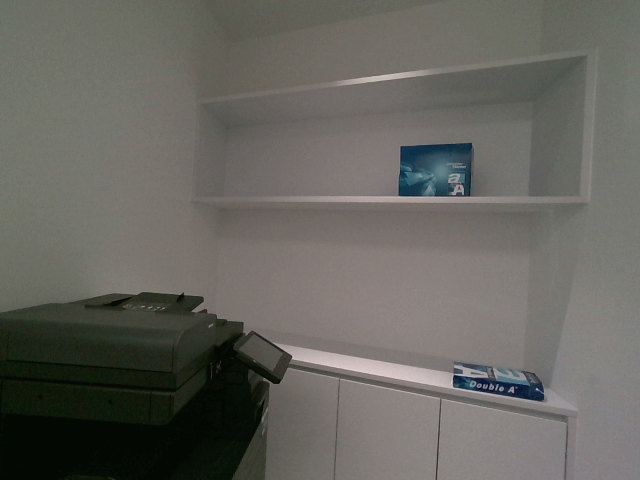} &
\includegraphics[width=0.23\textwidth]{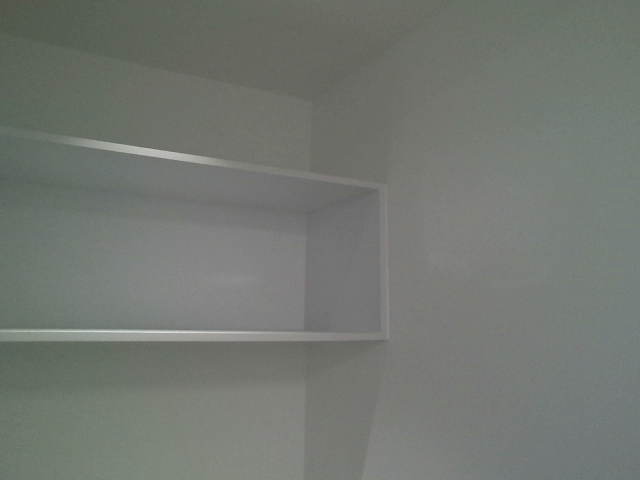} \\
(a) office\_01 & (b) office\_02 & (c) printer\_01 & (d) printer\_02 \\[6pt]
\includegraphics[width=0.23\textwidth]{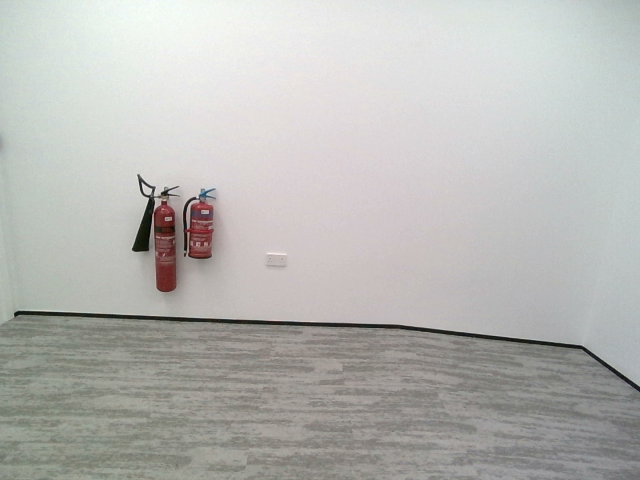} &
\includegraphics[width=0.23\textwidth]{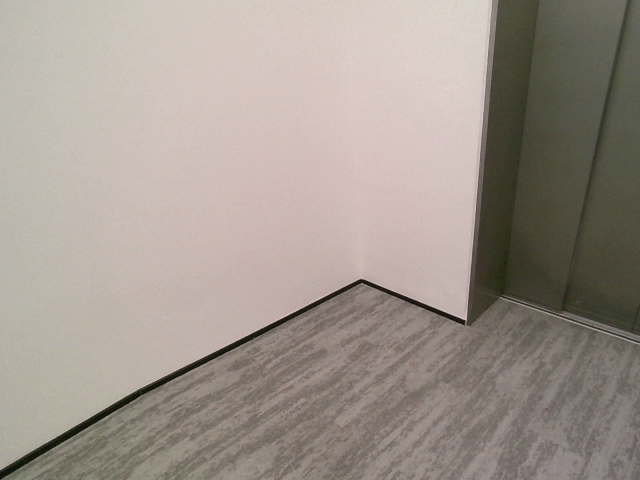} &
% \begin{tikzpicture}
% \node[inner sep=0pt] (img) {\includegraphics[width=0.23\textwidth]{figures/outdoor01.png}};
% \fill[white] (img.south west) rectangle (img.north east);
% \draw[black, thick] (img.south west) rectangle (img.north east);
% \node[black, font=\small\bfseries] at (img.center) {Anonymized};
% \end{tikzpicture} &
\includegraphics[width=0.23\textwidth]{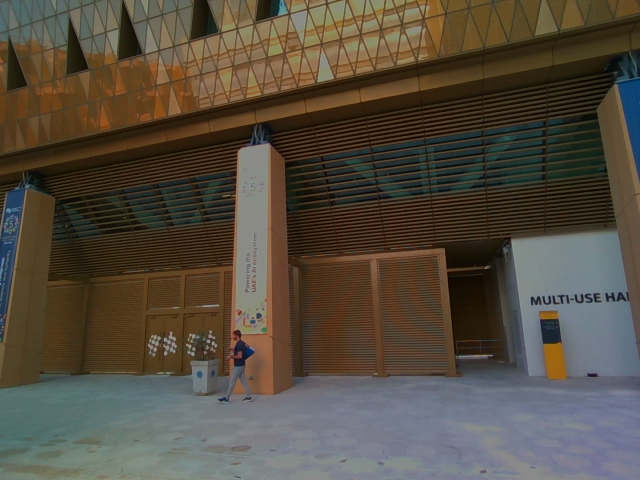} &
\includegraphics[width=0.23\textwidth]{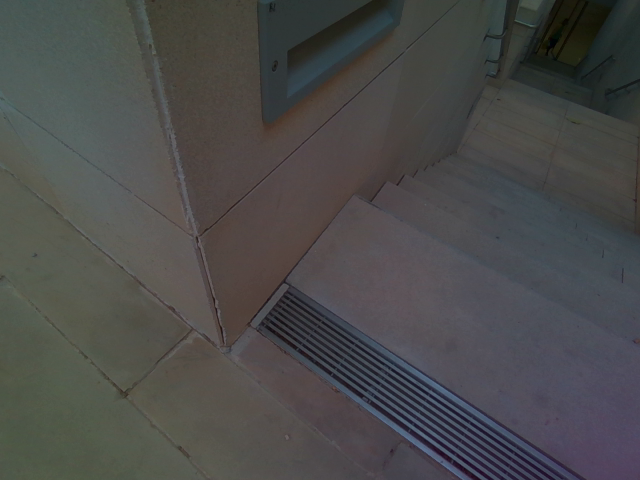} \\
(e) lift\_01 & (f) lift\_02 & (g) outdoor\_01 & (h) outdoor\_02 \\[6pt]
% & \begin{tikzpicture}
% \node[inner sep=0pt] (img) {\includegraphics[width=0.23\textwidth]{figures/outdoor03.png}};
% \fill[white] (img.south west) rectangle (img.north east);
% \draw[black, thick] (img.south west) rectangle (img.north east);
% \node[black, font=\small\bfseries] at (img.center) {Anonymized};
% \end{tikzpicture} &
& \includegraphics[width=0.23\textwidth]{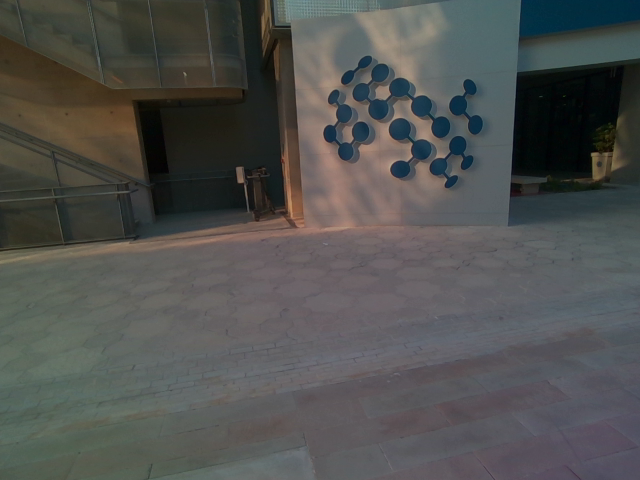} &
\includegraphics[width=0.23\textwidth]{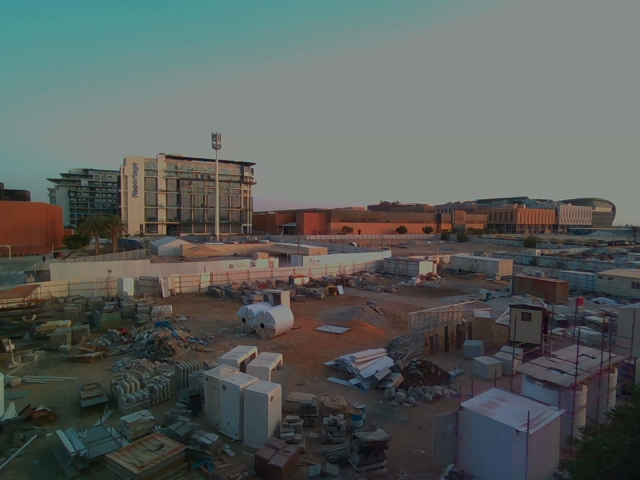} & \\
& (i) outdoor\_03 & (j) outdoor\_04 & \\
\end{tabular}
\caption{Sample frames from the self-collected dataset. (a)-(b) Indoor-Structured: office scenes with rich texture. (c)-(d) Indoor-Degraded: printer area with repetitive patterns. (e)-(f) Indoor-Degraded: indoor scenes with limited features. (g)-(j) Outdoor: diverse outdoor environments.}
\label{fig:dataset_samples}
\end{figure*}
\begin{table*}[!htbp]
\centering
\setlength{\tabcolsep}{10pt}
\caption{Ablation study using $\pi^3$ as the feed-forward 3D model.}
\label{tab:pi3_ablation}
\begin{tabular}{lcccccccc}
\toprule
& \makecell{Grav.\ Err.\\Lin. ($^\circ$) $\downarrow$} & \makecell{Vel.\ Err.\\Lin. (m/s) $\downarrow$} & \makecell{Scale Err.\\Lin.\ (\%) $\downarrow$} & \makecell{Scale Err.\\NL.\ (\%) $\downarrow$} & \makecell{Win.\ ATE\\($^\circ$/m) $\downarrow$} & \makecell{Traj.\ ATE\\($^\circ$/m) $\downarrow$} & \makecell{$T_{\mathrm{tot}}$\\(s) $\downarrow$} & \makecell{SR\\(\%) $\uparrow$} \\
\midrule
Pts.\ 20 & 2.04 & 0.28 & 35.48 & 23.13 & 2.22/0.046 & 1.54/0.038 & 1.07 & 96.0 \\
Pts.\ 50 & 2.02 & 0.26 & 29.32 & 18.90 & 2.23/0.047 & 1.65/0.042 & 1.07 & 92.0 \\
\textbf{Pts.\ 100} & \textbf{1.80} & \textbf{0.27} & \textbf{26.82} & \textbf{20.68} & \textbf{2.00/0.050} & \textbf{1.81/0.054} & \textbf{1.05} & \textbf{92.0} \\
Pts.\ 200 & 1.77 & 0.26 & 21.73 & 27.50 & 1.97/0.051 & 1.91/0.054 & 1.07 & 89.3 \\
Pts.\ 500 & 1.79 & 0.26 & 23.30 & 19.30 & 1.94/0.053 & 2.31/0.079 & 1.11 & 89.3 \\
Pts.\ 1000 & 1.40 & 0.18 & 25.67 & 19.70 & 1.50/0.043 & 2.07/0.060 & 2.15 & 93.3 \\
\midrule
w/o RANSAC & 1.85 & 0.27 & 35.90 & 31.30 & 1.99/0.048 & 1.75/0.053 & 1.06 & 90.7 \\
\midrule
Global Scale & 1.91 & 0.27 & 28.08 & 21.05 & 2.11/0.050 & 1.77/0.049 & 1.07 & 92.0 \\
\bottomrule
\end{tabular}
\end{table*}

\section{Self-Collected Dataset Details}
\label{sec:dataset}

As described in the main paper, we collect a dataset using a RealSense D455 camera, which provides RGB images at 30\,Hz and consumer-grade IMU measurements at 200\,Hz. Each sequence is approximately 25 seconds long and follows a figure-eight motion pattern to facilitate reliable initialization validation. The dataset spans three categories, as illustrated in Fig.~\ref{fig:dataset_samples}:
\paragraph{Indoor-Structured}
Indoor office environments with rich textures and complex depth structures.

\paragraph{Indoor-Degraded}
Visually challenging indoor environments, including scenes with repetitive textures and limited visual information.

\paragraph{Outdoor}
Outdoor scenes with varying lighting conditions, dynamic pedestrians, and distant or textureless regions.

\begin{figure*}[!htbp]
\centering
\includegraphics[width=\textwidth]{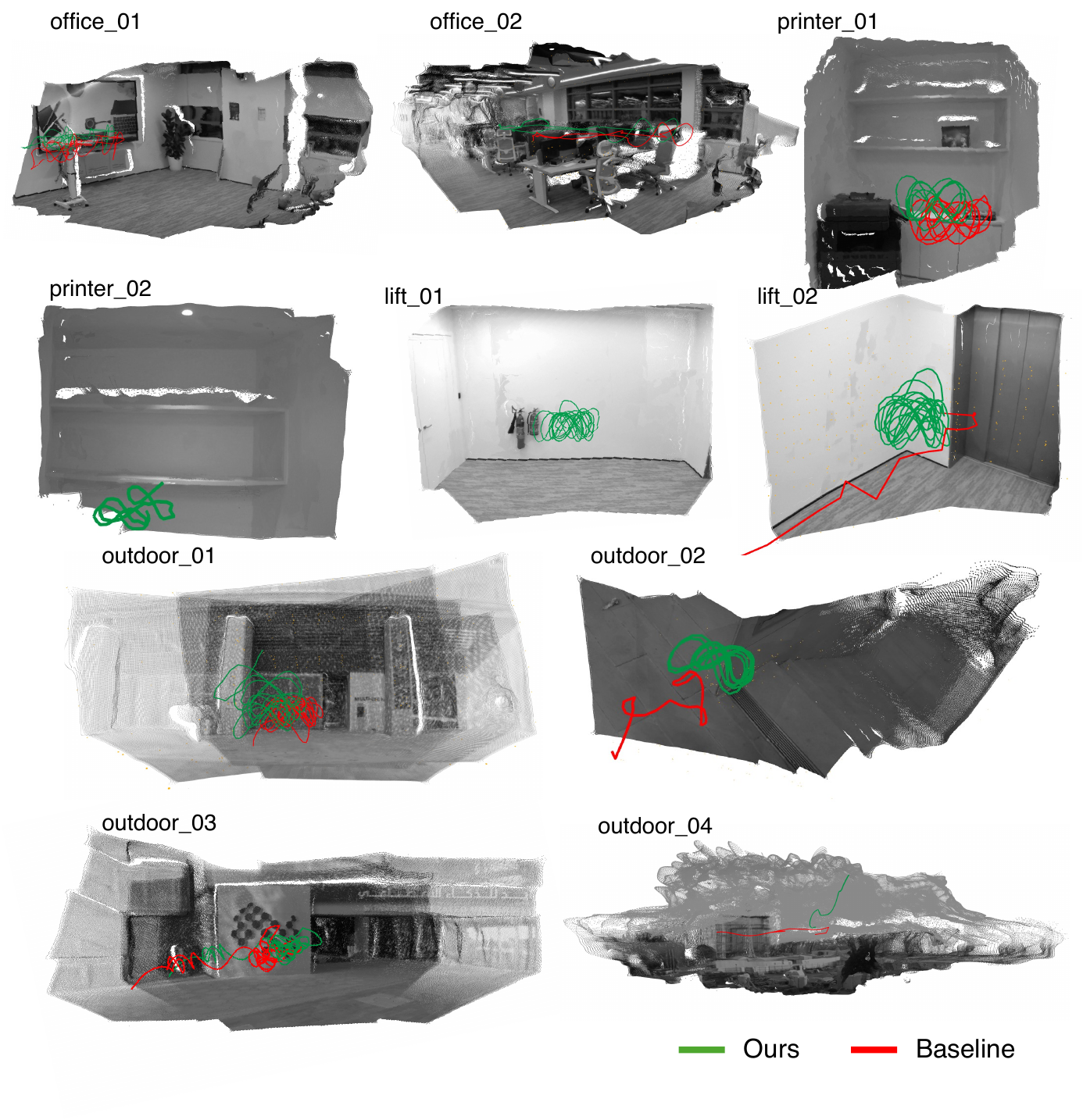}
\caption{Qualitative trajectory comparison on the self-collected dataset. Green: Ours. Red: Baseline.}
\label{fig:dataset_traj}
\end{figure*}
In \textit{Indoor-Structured} scenes, rich 2D visual information makes the setup favorable for traditional methods; however, as shown in Fig.~\ref{fig:dataset_samples}(b), the presence of complex depth variations is challenging for learning-based methods. Benefiting from the strong geometric inference capability of feed-forward 3D models, all learning-based methods remain effective under such conditions. 

In \textit{Indoor-Degraded} environments, the scenes are characterized by a small number of reliable visual measurements (typically 50–60 per frame) and repetitive textures. Under these conditions, feature-based baselines often fail to satisfy the initialization prerequisites of systems, or produce unreliable initial estimates due to insufficient geometric constraints, leading to obvious trajectory divergence even when VINS tracking is triggered. In contrast, the proposed feature-free formulation is less sensitive to sparse or repetitive visual patterns and demonstrates improved robustness.

\textit{Outdoor} scenes are particularly challenging for dynamic visual–inertial initialization. Large open areas and distant structures lead to limited parallax and weak depth variation within short temporal windows, posing a fundamental difficulty shared by all approaches. These challenges are further amplified when relying on feed-forward 3D models, as outdoor environments often contain textureless or ambiguous regions (e.g., open sky or reflective surfaces) and exhibit a wide range of scene depths, resulting in unreliable predictions. While successful initialization is still achievable in some outdoor scenarios, extreme cases such as the open-sky setting in \textit{outdoor\_04} remain harsh for learning-based methods.

\end{document}